\pdfoutput=1
\documentclass[11pt]{article}
\usepackage{acl2023}
\usepackage{times}
\usepackage{latexsym}
\usepackage[T1]{fontenc}
\usepackage[utf8]{inputenc}
\usepackage{microtype}
\usepackage{inconsolata}
\usepackage[utf8]{inputenc} 
\usepackage[T1]{fontenc}    
\usepackage{hyperref}       
\usepackage{url}            
\usepackage{booktabs}       
\usepackage{amsfonts}       
\usepackage{nicefrac}       
\usepackage{microtype}      
\usepackage{amssymb}
\usepackage{enumitem}
\usepackage{times}
\usepackage{epsfig}
\usepackage{graphicx}
\usepackage{amsmath}
\usepackage{xcolor}
\usepackage{pdftexcmds}
\usepackage{xspace}
\usepackage{dsfont}
\usepackage[normalem]{ulem}
\usepackage{hyperref}
\usepackage{url}
\usepackage{multirow}
\usepackage{subfig}
\usepackage{caption} 
\usepackage{wrapfig}
\usepackage{titlesec}
\definecolor{color5}{HTML}{006795}
\definecolor{lightgray}{gray}{0.75}
\hypersetup{
  colorlinks   = true, 
  urlcolor     = color5, 
  linkcolor    = color5, 
  citecolor   = color5 
}

\title{The Unreasonable Effectiveness of Easy Training Data for Hard Tasks}

\author{
 \textbf{Peter Hase}$^{1,2}$  \ \ \ \ \ \ \
 \textbf{Mohit Bansal}$^{2}$  \ \ \ \ \ \ \ \
 \textbf{Peter Clark}$^{1}$  \ \ \ \ \ \
 \textbf{Sarah Wiegreffe}$^{1}$ \\
 $^{1}$Allen Institute for AI~~~~~~
 $^{2}$UNC Chapel Hill \\
 \small\texttt{ \{peter, mbansal\}@cs.unc.edu}, 
 \small\texttt{ peterc@allenai.org},
 \small\texttt{ wiegreffesarah@gmail.com}
}

\begin{document}
\maketitle
\begin{abstract}
\looseness=-1
How can we train models to perform well on hard test data when hard training data is by definition difficult to label correctly?
This question has been termed the \emph{scalable oversight} problem and has drawn increasing attention as language models have continually improved.
In this paper, we present the surprising conclusion that current pretrained language models often generalize relatively well from easy to hard data, even performing as well as oracle models finetuned on hard data.
We demonstrate this kind of easy-to-hard generalization using simple finetuning methods like in-context learning, linear classifier heads, and QLoRA for seven different measures of datapoint hardness, including six empirically diverse human hardness measures 
(like grade level) and one model-based measure (loss-based).
Furthermore, we show that even if one cares most about model performance on hard data, it can be better to collect easy data rather than hard data for finetuning, since hard data is generally noisier and costlier to collect. 
Our experiments use open models up to 70b in size and four publicly available question-answering datasets with questions ranging in difficulty from 3rd grade science questions to college level STEM questions and general-knowledge trivia.
We conclude that easy-to-hard generalization in LMs is surprisingly strong for the tasks studied.\footnote{Our code is publicly available at: \url{https://github.com/allenai/easy-to-hard-generalization}.}

\end{abstract}

\section{Introduction}

\begin{figure}[!t]
   \centering
   \includegraphics[width=.49\textwidth]{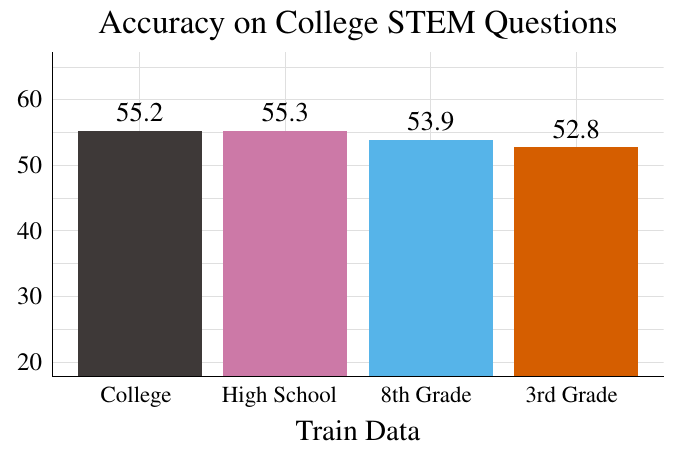}
   \vspace{-18pt}
   \caption{
   A model prompted with easy data (e.g., 3rd Grade problems) does \emph{almost as well} on a hard task (College problems) as a model prompted with hard data (the College bar).
   Results shown for Mixtral-8x7B with $k{=}10$ prompt examples, averaged over 5 random seeds.
   }
   \vspace{-5pt}
   \label{fig:rq0_third_grade_to_college}
\end{figure}

\looseness=-1
It is difficult to supervise LMs (i.e., train LMs to give correct outputs) in specialized domains of human knowledge, because it is difficult to correctly label data in such domains. Labeling difficulty manifests itself in both time to annotate (and thus cost) and label noise \cite{lease2011quality, northcutt2021pervasive}. 
Labeling difficulty becomes severe when specific expertise is required \cite{sambasivan2021everyone}. For example, for sufficiently specific physics problems, PhD holders and PhD students can make errors on as many as 40\% of (objective) problems \cite{rein2023gpqa}. 
As more NLP benchmarks focus on challenging domain-specific tasks, having access to large human-labeled training corpora may become increasingly infeasible (e.g., existing benchmarks like MMLU \cite{hendrycks2020measuring} and GPQA \cite{rein2023gpqa} do not come with training data). The question arises: how can we train models to solve hard problems when correctly labeling enough hard data for training is difficult? 
This problem is an example of the scalable oversight problem, which concerns how to give a good reward signal to a model when it is difficult to assess if its outputs are correct \cite{amodei2016concrete}. 

\begin{figure*}[!ht]
   \centering
   \includegraphics[width=.98\textwidth]{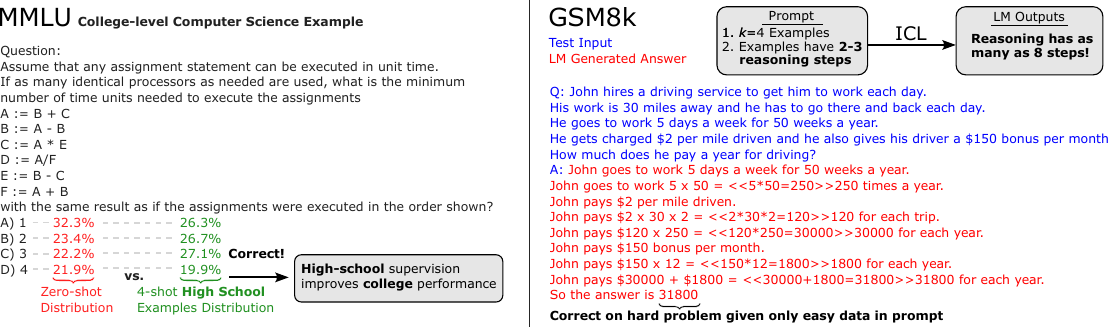}
   \vspace{-6pt}
   \caption{Supervising Llama-2-70b with \emph{easy} data (left: high school level computer science problems; right: math problems with 2-3 reasoning steps) can enable generalization to \emph{hard} data (left: a college level computer science problem; right: a math problem involving 8 reasoning steps). Prompts themselves are omitted for brevity.
   }
   \vspace{-2pt}
   \label{fig:examples-fig}
\end{figure*}

\looseness=-1
In this paper, we study the problem of \textbf{easy-to-hard generalization}.
Easy-to-hard generalization refers to model performance on hard test data when finetuned\footnote{We use ``finetuning'' interchangeably with ``training'' and ``fitting'' to refer to fitting pretrained models to data via in-context learning (ICL), parameter efficient finetuning (QLoRA), or by training a linear classifier head.} only on easy training data, defined according to some human hardness measure (like grade level). Since gathering data in domains like graduate level STEM fields is expensive and time-consuming, it would clearly be useful if we could improve model performance in these domains by only finetuning models on cleanly labeled data from simpler domains, like high school STEM questions. To assess how well current LMs generalize from easy to hard data, we fit models to easy data and test them on hard data (``easy-to-hard''), then compare them to an oracle upper bound and unsupervised lower bound. The oracle upper bound is a model that has access to labeled hard data for finetuning (``hard-to-hard''), while the unsupervised lower bound is a model that is prompted zero-shot to answer questions (``unsupervised-to-hard'').
The metric we are interested in is the \textbf{Supervision Gap Recovered (SGR):}
\begin{align*}
    \frac{\textrm{Easy} - \textrm{Unsupervised}}{\textrm{Hard} - \textrm{Unsupervised}}
\end{align*}
\hspace{-5pt} where Easy, Hard, and Unsupervised refer to model performance \emph{on hard test data} when finetuned on easy data, hard data, or no data (zero-shot), respectively.
This metric takes a value of 100\% when finetuning on easy data is as effective as hard data, and it is 0\% when a model finetuned on easy data is no better than prompting a model zero-shot.

Our main result is that pretrained language models generalize surprisingly well from easy to hard data, often performing almost as well as an oracle model finetuned on hard data (illustrated in Fig. \ref{fig:rq0_third_grade_to_college}).
In experiments with ARC \cite{Clark2018ThinkYH}, MMLU \cite{hendrycks2020measuring}, GSM8k \cite{cobbe2021training}, and StrategyQA \cite{Geva2021DidAU}, we find that the Supervision Gap Recovered is usually \textbf{between~70\%~and 100\%}, meaning that easy supervision is at least 70\% as good as hard supervision for hard test performance (see Fig. \ref{fig:examples-fig} for example problems).
These results are robust across (1) model family and scale (between 7b and 70b parameters), (2) six different human hardness measures and a model-based measure, (3) four datasets/tasks, and (4) several finetuning methods including in-context learning with and without chain-of-thought reasoning \cite{brown2020language, wei2022chain}, QLoRA \cite{dettmers2023qlora}, and linear classifier heads \cite{belinkov2022probing}.
Overall, our results suggest that current LMs generalize relatively well to test data across human difficulty levels even when finetuned on data that is measurably easier than the test data. We hypothesize that this occurs because easy data elicits latent knowledge and skills from pretrained models \emph{in a hardness-invariant way}.

We additionally demonstrate that easy supervision can outperform hard supervision when (1) within some data collection budget, a greater quantity of easy data can be collected than hard data, or (2) easy data can be labeled with lower error rates than hard data (Sec. \ref{sec:RQ3}). 
Lastly, we study how easy-to-hard generalization changes with model scale and the gap between train and test hardness (Sec. \ref{sec:RQ4}). 
The remainder of the paper is organized along the following research questions:
\vspace{-3pt}
\begin{enumerate}[leftmargin=31pt, itemsep=-3pt, label=\textbf{RQ\arabic*:}]
    \item How Can We Measure Data Hardness? 
    \item Can We Do Well on Hard Data by Training on Easy Data?
    \item What Are the Cost-Benefit Tradeoffs of Collecting Easy vs. Hard Training Data? 
    \item Is Easy-To-Hard Generalization Consistent Across Model Scale and Train-Test Hardness Gap Size?
    \vspace{-3pt}
\end{enumerate}
We summarize our main conclusions below: 
\vspace{-3pt}
\begin{enumerate}[leftmargin=12pt, itemsep=-2pt]
    \item Our six human hardness measures and one model-based measure are empirically diverse and model performance declines on harder test data for each measure.
    \item LMs generalize surprisingly well from easy-to-hard data, closing 70\%-100\% of the gap between unsupervised and hard train data.
    \item We show that it is often better to train on easy data when hard data is more expensive to collect or has noisier labels. 
    \item The Supervision Gap Recovered is highly robust across model scale.  Easy-to-hard performance may begin to decline when the train-test hardness gap is large enough. 
\end{enumerate}

\section{Related Work}

\noindent \textbf{Curriculum Learning}. Curriculum learning has historically concerned itself with model performance on hard data points \cite{bengio2009curriculum}. Previous work in this area has argued that learning from easy data first is helpful for models later learning more complex concepts and therefore performing better on hard data \cite{xu2020curriculum}. 
Whereas curriculum learning aims to improve hard test performance by optimally ordering the training data, we simply aim to investigate how well models generalize to hard data based on the hardness of the training data.
Our results suggest that pretrained LMs generalize surprisingly well from easy to hard data, potentially alleviating the need for heavily engineered training curricula.

\looseness=-1
\vspace{4pt}
\noindent \textbf{Compositional Generalization}. 
Work in compositional generalization has previously shown that neural networks struggle to generalize to problems that require combining reasoning steps in ways not seen exactly during training \cite{lake2018generalization, bogin2022unobserved, zhou2023data}. 
Further work has begun characterizing the conditions under which models generalize to compositionally more difficult problems. For instance, Transformers will generalize better on classes of algorithmic problems whose solutions can be written in RASP, meaning the programs can be implemented exactly by a Transformer forward pass \cite{zhou2023algorithms}. 
Recurrent test-time computation also appears to be quite valuable for generalizing to problems requiring more reasoning steps than those seen during training \cite{schwarzschild2021can, bansal2022end}. 
Interestingly, even GPT-3.5 with {Chain-of-Thought} prompting can struggle to generalize to simple mathematical problems requiring more reasoning steps than seen during finetuning \cite{dziri2023faith}. Our results are not inconsistent with these studies, but instead demonstrate that \emph{relative to} an unsupervised-to-hard lower bound and hard-to-hard upper bound, easy-to-hard performance on compositional reasoning problems is often surprisingly good (Sec. \ref{sec:RQ2}). 

\vspace{3pt}
\noindent \textbf{Easy-to-Hard Generalization}. 
\citet{amodei2016concrete} motivate the scalable oversight problem by pointing out how it could be challenging to give a proper reward signal to a model when it is difficult to assess if its outputs are correct. Assessing easy-to-hard generalization provides useful context for understanding the difficulty of the scalable oversight problem, as it tells us how we can expect models to generalize from a setting where we can properly supervise them to one where we cannot.
Past work evaluates easy-to-hard generalization in NLP using model-based hardness measures \cite{swayamdipta2020dataset} and number of compositional reasoning steps required to solve a problem \cite{fu2022complexity}.
\citet{swayamdipta2020dataset} show that BERT models perform worse on commonsense reasoning tasks when finetuned on easy data rather than hard data according to a loss-based metric resembling minimum description length \cite{perez2021rissanen}. \citet{fu2022complexity} show a similar result with GPT3 models for StrategyQA and GSM8k, finding that prompting with ``complex'' examples does better than ``simple'' examples, where examples are categorized according to the number of reasoning steps in the annotated human reasoning chain for a problem. 
Relative to these works, we study easy-to-hard generalization with (1) a greater number of human hardness measures, including grade level, expert rating, required cognitive skills, question length, answer length, and number of reasoning steps, as well as a model-based metric, (2) multiple datasets involving science question answering, compositional reasoning, and mathematical reasoning, and (3) multiple model sizes for understanding scaling trends. In contrast to these works, we show that in a number of settings easy-to-hard generalization is comparable to hard-to-hard generalization. 

\looseness=-1
In concurrent work, \citet{burns2023weak} present results on a related ``weak-to-strong'' generalization problem, where a stronger model is finetuned using labels from a weaker model. They also explore easy-to-hard generalization for NLP tasks using a model-based hardness measure. In contrast to this work, (1)~we define our main performance metric (Supervised Gap Recovered) using an unsupervised model as the baseline performance rather than a \emph{weaker} model as the baseline performance, which is important when an unsupervised stronger model will greatly outperform a supervised weaker model (as is observed in our experiments); (2)~we use human hardness measures in addition to model-based hardness, which is a more realistic and important setting when the two may not correlate strongly (see our Fig.~\ref{fig:rq1_appendix}); and (3)~we use publicly available datasets and open-source models rather than unidentified ``NLP tasks'' and API-gated models.

\begin{table*}[th]
\setlength{\tabcolsep}{11pt} 
\centering
\resizebox{\textwidth}{!}{
\small
\begin{tabular}{l l l l}
\toprule
\textbf{ARC} & \textbf{MMLU-STEM-5} & \textbf{StrategyQA} & \textbf{GSM8k} \\
\midrule
$n=4521$ & $n=1746$ & $n=2290$ & $n=8792$ \\
\midrule
Grade Level (3-8) & Grade Level (HS vs. College) & \textcolor{lightgray}{Grade Level} & \textcolor{lightgray}{Grade Level} \\
Difficulty Score (1-3) & \textcolor{lightgray}{Difficulty Score} & \textcolor{lightgray}{Difficulty Score} & \textcolor{lightgray}{Difficulty Score} \\
Bloom Skill (1-5) & \textcolor{lightgray}{Bloom Skill} & \textcolor{lightgray}{Bloom Skill} & \textcolor{lightgray}{Bloom Skill} \\
Question Num. Words & Question Num. Words & Question Num. Words & Question Num. Words \\
Answer Num. Chars & Answer Num. Chars & Answer Num. Chars & Answer Num. Chars \\
\textcolor{lightgray}{Num. Reasoning Steps} & \textcolor{lightgray}{Num. Reasoning Steps} & Num. Reasoning Steps & Num. Reasoning Steps \\
MDL & MDL & MDL & MDL \\
\bottomrule
\end{tabular}
}
\vspace{-5pt}
\caption{Hardness measures we use for each dataset. Grayed-out options are not present in the dataset's annotations, and thus not used in our experiments.}
\label{tab:hardness_vars}
\end{table*}

\section{Measuring Datapoint Hardness}

\looseness=-1
Measuring easy-to-hard generalization requires drawing a distinction between easy and hard data, defined in terms of human ability to correctly label the data. 
There could be many ways to describe what makes problems harder, including that (1) only people with specialized training and knowledge can solve the problem \cite{lehman-etal-2019-inferring}; (2) it takes people longer to solve the problem; (3) people are less certain that their final solution is correct; (4) people with similar expertise naturally disagree about the solution to the problem, while agreeing that there is an objective solution \cite{dumitrache2018crowdsourcing, pavlick2019inherent, nie-etal-2020-learn}; (5) experts know of a reliable method for obtaining the answer to a problem, but it is costly in terms of time and effort or possibly noisy in its outputs (like conducting scientific experiments).
In this paper, we aim to capture the above properties in a number of specific measures we can obtain for each instance in our datasets, including:
\begin{enumerate}[leftmargin=12pt, itemsep=-1pt]
\item \textbf{Education/Grade Level:} What education level (possibly in a particular domain) would typically lead one to be able to answer the question?
\item \textbf{Expert Rating:} How difficult would an expert rate the question, on an ordinal scale?
\item \textbf{Required Cognitive Skill:} What cognitive skills are required by the question? This rating is based on Bloom's cognitive skills taxonomy, in order of increasing complexity: (1) Remembering, (2) Understanding, (3) Applying, (4) Analyzing, and (5) Evaluating \cite{bloom1956taxonomy, adams2015bloom}.
\item \textbf{Question Num. Words:} Question length is a natural proxy for question hardness, as longer questions can involve more premises or a greater number of concepts. 
\item \textbf{Answer Num. Chars:} We also consider Answer Num. Chars, since longer answers may reflect more specific or more complex problems. Character count provides a measure that is applicable across tasks.

\item \textbf{Compositional Steps:} Compositional reasoning is more difficult than executing individual reasoning ``primitives.'' 
We consider how many individual reasoning steps are involved in answering a question (i.e., the number of subproblems whose solutions must be combined), according to human-annotated reasoning chains. 

\item \textbf{Minimum Description Length:} A model-based measure of hardness, measuring datapoint loss under a model family. Details for MDL computation are given in Appendix \ref{app:MDL}.

\end{enumerate}

\looseness=-1
\noindent The first six hardness measures are fundamentally human notions of hardness, but we can also measure a model-based metric for datapoint hardness. In this direction, the seventh measure is a minimum-description-length (MDL) metric \cite{voita2020information}. In practice, MDL can be measured by computing a \emph{test} datapoint's average label probability across models of identical architecture finetuned on increasing quantities of training data for a task \cite{perez2021rissanen}. Intuitively, MDL captures how hard on average an in-distribution test datapoint is for a model to generalize to given some amount of training data.
Ultimately, we use our MDL metric to capture how well a \emph{stronger} model generalizes to data that is hard according to a \emph{weaker} model, in order to simulate a setting where humans cannot label hard problems that they would like for a strong model to solve.

In our experiments, we use four datasets that contain instance-level annotations for some portion of these measures, as shown in Table \ref{tab:hardness_vars}. 

\begin{enumerate}[leftmargin=12pt, itemsep=-1pt, label=$\bullet$]
\item \textbf{ARC} \cite{Clark2018ThinkYH}: U.S. gradeschool science questions in multiple-choice format. We combine ARC-Easy and ARC-Challenge splits. Random performance is 25\%.
\item \textbf{MMLU} \cite{hendrycks2020measuring}: Domain-specific multiple-choice questions for many domains. We subset to high school and college level math, physics, biology, chemistry, and computer science questions (MMLU-STEM-5). Grade level is high school (HS) vs. college. See \autoref{fig:examples-fig} (left) for an example. Random performance is 25\%.
\item \textbf{StrategyQA} \cite{Geva2021DidAU}: Yes/no general knowledge trivia questions requiring compositional reasoning over individual facts. The ``Num. Reasoning Steps'' measure is the number of facts that must be combined. Majority-class vote performance is 53.9\%.
\item \textbf{GSM8k} \cite{cobbe2021training}: U.S. grade school math word problems in direct answer format (i.e., no answer choices given). Random performance is 0\%. The number of steps in a problem solution is the ``Num. Reasoning Steps'' measure and is obtained from the human-annotated reasoning chain collected for each problem. 
See \autoref{fig:examples-fig} (right) for an example.
\end{enumerate}

\noindent There are generally fewer hard datapoints than easy datapoints in our datasets, given the relative difficulty of collecting hard data. In MMLU-STEM-5, for example, there are 603 college level questions and 1143 high school questions. We show histograms for each hardness measure distribution in Appendix Fig. \ref{fig:hardness_var_distr}. See Appendix \ref{app:dataset_details} for further dataset information.

\section{Experiment Setup}
\label{sec:experiment_setup}

\textbf{Models.} Apart from Fig. \ref{fig:rq0_third_grade_to_college}, we report results in the main paper on the Llama-2 70b base model \cite{touvron2023llama2}. In Appendix \ref{app:additional_results}, we show results for Llama-2 7b and 13b, an RLHF version of Llama-2-70b (``Llama-2-chat''), Qwen-72b \cite{qwen}, and Mixtral-8x7b \cite{jiang2024mixtral}. 

\vspace{4pt}
\looseness=-1
\noindent\textbf{Data Hardness Stratification.} To separate datasets into easy and hard data (with leftover data being medium data), we define easy/hard cutoffs as follows: for Question Num. Words, Answer Num. Chars, and MDL, we automatically define these values to be at the 30th and 70th percentiles of the variable range. Other variable cutoffs are defined manually: For \textbf{ARC}, \emph{Grade Level} is easy (3-5), medium (6-7), hard (8); \emph{Difficulty Score} is easy (1), medium (2), hard (3); \emph{Bloom Skill} is easy (1-2), medium (3), hard (4-5). For \textbf{MMLU}, \emph{Grade Level} is easy (high school) and hard (college) with no medium. For \textbf{StrategyQA}, \emph{Num. Reasoning Steps} is easy (1-2), medium (3), hard (4-5). For \textbf{GSM8k}, \emph{Num. Reasoning Steps} is easy (2-3), medium (4-5), hard (6-11). We show histograms for each hardness measure distribution in Appendix Fig. \ref{fig:hardness_var_distr}.

\looseness=-1
\vspace{4pt}
\noindent\textbf{Finetuning Methods.} We fit models to data with in-context learning \cite[ICL;][]{brown2020language}, linear classifiers trained on frozen model hidden states \cite{belinkov2022probing}, or QLoRA \cite{dettmers2023qlora}. StrategyQA and GSM8k benefit heavily from utilizing chain-of-thought reasoning \cite[CoT;][]{wei2022chain}, so we primarily conduct experiments for these datasets with ICL+CoT and QLoRA+CoT (using reasoning chains from the datasets for supervision). See descriptions of each method below, with full detail in Appendix \ref{app:tuning_details}. 

\looseness=-1
\begin{enumerate}[leftmargin=12pt, itemsep=-1pt]
\item \textbf{ICL}: For in-context learning (ICL), we use $k{=}10$ prompt examples for ARC and MMLU and $k{=}8$ examples for StrategyQA and GSM8k (we see diminishing returns for larger $k$). When scoring multiple choice questions (no CoT), we get a model prediction by computing the answer probability for each answer choice given the test input and the prompt. When generating outputs with CoT, we greedily generate up to $t=100$ tokens for StrategyQA and $t=300$ tokens for GSM8k. Accuracy is computed as exact match between predicted answer and label.

\item \textbf{Linear Probing}: We train a linear classifier on frozen LM hidden states. This is an effective method for performing multiple choice QA using LM representations \cite{liu2023cognitive}, and it does not require any finetuning of the underlying LM. 
For a given question, we compute one representation per answer choice by concatenating the question and answer choice as input and extracting the model's final-token representation.
Then, we score each representation $z$ by applying the linear probe: $f(z;w) = w^Tz$. The answer choice with the highest score is returned as the prediction. The probe weight $w$ is trained using SGD to minimize cross-entropy loss on a dataset of representations $Z = \{\{z_{i,j}\}_{j=1}^{j=|A|}\}_{i=1}^{N}$ derived from $N$ training datapoints with $|A|$ answer choices. 

\item \textbf{QLoRA}: 
To finetune our LMs, we execute QLoRA with the LoRA implementation from HuggingFace \texttt{peft} \cite{peft} and the 8-bit AdamW from \texttt{bitsandbytes} \cite{dettmers2022optimizers}. 
We train the default layers for Llama-2 with rank $r=16$ adapters, $\alpha=32$, and dropout $p=0.1$.
Model predictions are obtained in the same manner as for ICL, i.e., by scoring multiple choice options or generating $t=100/300$ tokens for StrategyQA/GSM8k.
\end{enumerate}

\noindent\textbf{Unsupervised Baseline.} Our unsupervised baseline is zero-shot prompting, scoring the answer choice probabilities given the question and taking the highest probability answer as the model prediction. The one exception to this is for GSM8k, which does not have multiple answer choices per question. For this dataset, we use a simple ``Let's think step by step'' style prompt. See Appendix \ref{app:tuning_details}.

\vspace{4pt}
\looseness=-1
\noindent\textbf{Training Size Controls.} For all experiments with linear probing and QLoRA, we use $n=160$ train points. While we would prefer to use more finetuning data, the bottleneck we face is that fairly comparing easy-to-hard with hard-to-hard generalization requires both fixing the amount of finetuning data and leaving enough hard data left over for testing. Since we have as few as $n=603$ hard test points for MMLU, we have to limit finetuning data to $n=160$ points to leave enough test data for reasonably small confidence intervals.

\vspace{4pt}
\looseness=-1
\noindent\textbf{Statistical Testing.} We perform experiments using 5 random seeds, controlling the training data selection (leaving remaining data for testing). 
To obtain confidence intervals and $p$-values, we use block bootstrap sampling \cite{efron1994introduction}. See Appendix \ref{app:statistical_testing} for further detail.

\section{Experiments}
\label{sec:experiments}

\begin{figure}[t]
   \centering
   \includegraphics[width=.46\textwidth]{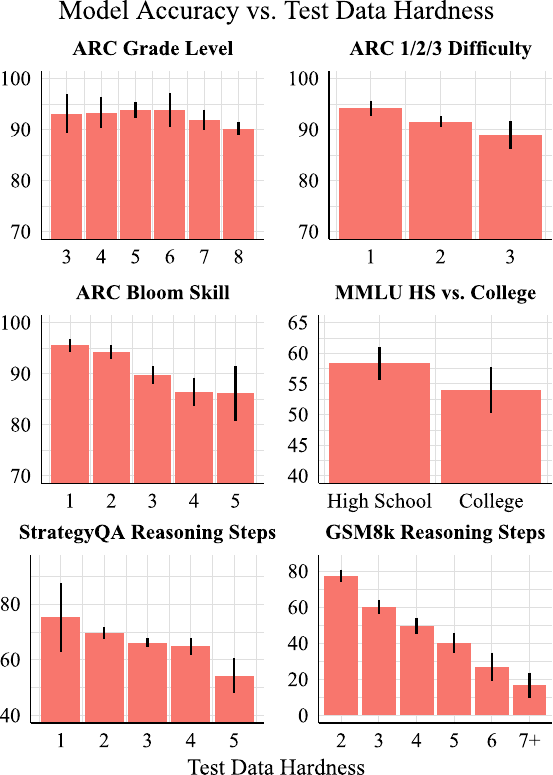}
   \vspace{-9pt}
   \caption{Accuracy vs test data hardness across datasets (using Llama-2-70b with ICL). Data that humans find harder is also harder for LMs. Error bars are 95\% CIs showing test sample variance.}
   \vspace{-5pt}
   \label{fig:rq2_acc_vs_test_hardness}
\end{figure}

\begin{figure*}[!ht]
   \centering
   \hspace{-8pt}
   \includegraphics[width=.92\textwidth]{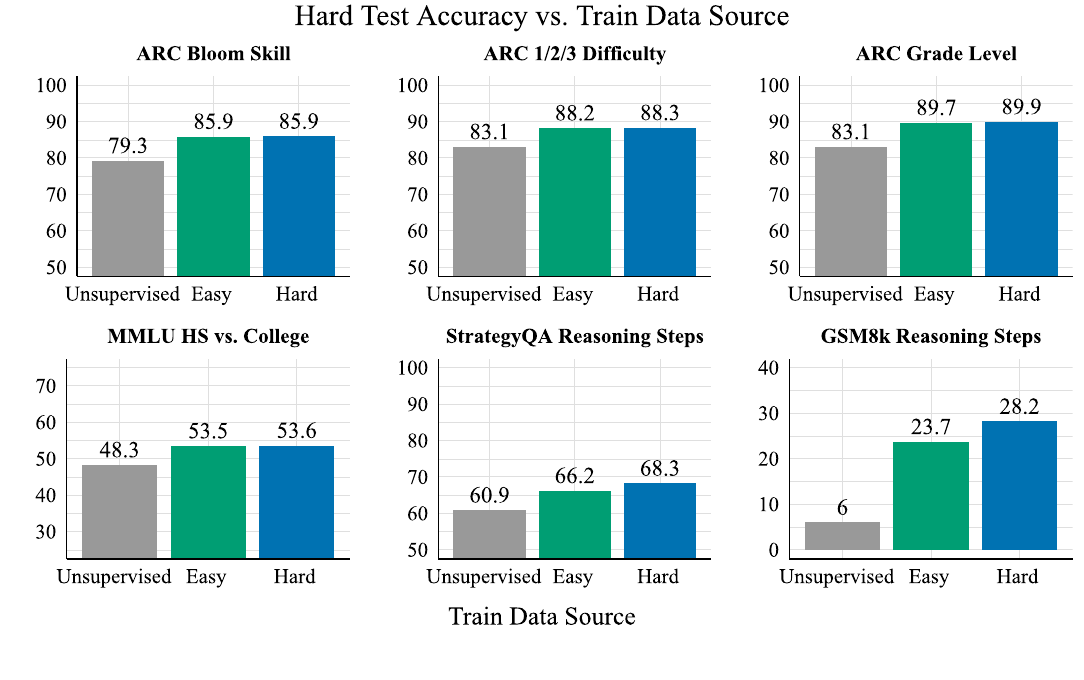}
   \vspace{-17pt}
   \caption{Accuracy on hard test data vs train data hardness (using Llama-2-70b and ICL, averaged over 5 seeds). Models recover 70-100\% of the supervision gap (between Unsupervised and Hard) when finetuned on Easy data.}
   \label{fig:rq3_acc_vs_train_hardness}
\end{figure*}

\subsection{RQ1: How Can We Measure Data Hardness?}
\label{sec:RQ1}

We first explore properties of our hardness measures for each dataset. Here, we focus on hardness measures unique to each dataset, with full results across all measures shown in Appendix \ref{app:additional_results}.

\vspace{4pt}
\noindent \textbf{Design.} While our human hardness measures are direct measurements of data hardness, we validate that each measure is meaningful by assessing model performance across test hardness levels, using Llama-2-70b and ICL with randomly sampled prompts.
We also create correlation heatmaps for hardness measures in our datasets, using a Spearman rank-order correlation \cite{spearman1987proof} between hardness values for each datapoint. 

\vspace{5pt}
\noindent \textbf{Results.} It appears that all of our hardness measures meaningfully capture some aspect of datapoint hardness, as model accuracy declines for harder test data for each of these measures (Fig. \ref{fig:rq2_acc_vs_test_hardness}), including for model-based hardness as measured by an ensemble of 7b-parameter models (Appendix Fig. \ref{fig:rq2_appendix}). This also holds when finetuning with QLoRA (Appendix Fig. \ref{fig:rq2_qlora}).
Next, we find that our hardness measures are empirically very diverse (Appendix Fig. \ref{fig:rq1_appendix}). Correlations between hardness measures are fairly low, suggesting that these measures capture different possible aspects of datapoint hardness. 
We conclude that easy-to-hard generalization should be assessed with multiple notions of datapoint hardness, since there are several different available measures and model performance declines for harder test data along each measure.

\subsection{RQ2: Can We Do Well on Hard Data by Training on Easy Data?}
\label{sec:RQ2}

We now examine how well models generalize from easy training data to hard test data.

\vspace{4pt}
\noindent \textbf{Design.} For each of our hardness measures, we test models on exclusively hard test data (according to that hardness measure), while varying whether they are finetuned on easy or hard data.\footnote{We report test accuracy on the full data distribution and the easy test split in Appendix Figs. \ref{fig:RQ3_all_acc} and \ref{fig:RQ3_easy_acc}, respectively.}

\vspace{5pt}
\noindent \textbf{Results.} Surprisingly, \textbf{Llama-2-70b with ICL shows comparable generalization to hard test data regardless of whether it is fit to easy or hard data} (Fig.~\ref{fig:rq3_acc_vs_train_hardness}). 
In fact, across all six hardness measures, the \textbf{Supervision Gap Recovered is between 70\% and 100\%}. These results are statistically significant, with CIs and $p$-values shown in Appendix Table \ref{tab:RQ3-p-values}.
Interestingly, for ARC and MMLU, there is \emph{no difference} in easy vs. hard generalization using ICL.
Results are also robust across finetuning methods and additional hardness measures (Appendix Figs. \ref{fig:RQ3_SGR_by_methods_main}, \ref{fig:RQ3_appendix}). 
With QLoRA, for example, the SGR remains within 70\%-100\% for ARC, MMLU and StrategyQA.
While GSM8k appears to exhibit worse easy-to-hard generalization, we note that easy-to-\emph{all} generalization is actually equally good to hard-to-\emph{all} generalization (see Fig. \ref{fig:RQ3_all_acc}).
Thus it seems like easy data provides surprisingly good supervision for LMs.

\looseness=-1
These results contrast notably with past work in curriculum learning and compositional generalization \cite{bengio2009curriculum, lake2018generalization}.
This is likely because models like Llama-2-70b have learned much more during pretraining than models commonly used in work on curriculum learning and compositional generalization. So, it would seem that finetuning these models on relatively small amounts of easy data successfully elicits the relevant task knowledge from the models in a way that is largely invariant to datapoint hardness.

\subsection{RQ3: What Are the Cost-Benefit Tradeoffs of Collecting Easy vs. Hard Training Data?}
\label{sec:RQ3}

\looseness=-1
One implication of the results from Sec. \ref{sec:RQ2} is that if easy data is almost as good as hard data, it could be better to collect and fit to easy data, since hard data can be noisier and costlier to collect \cite{sambasivan2021everyone}.
Hence, we test the hypothesis that finetuning on easy data outperforms hard data under two possible assumptions: (1) that one can collect more easy data than hard data given a fixed budget (i.e., time, money), and (2) that easy data is less noisily labeled than hard data. 

\begin{figure}[!t]
   \centering
   \hspace{-1pt}
   \includegraphics[width=.48\textwidth]{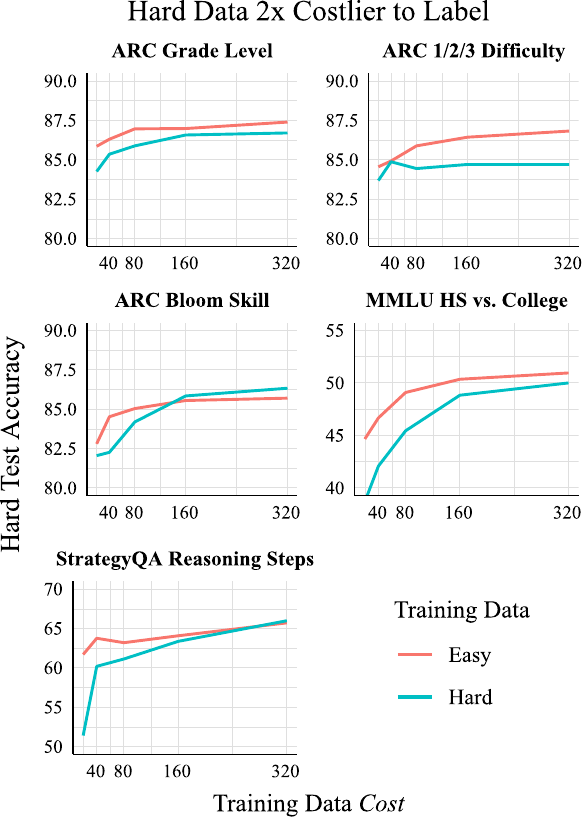}
   \vspace{-14pt}
   \caption{Hard test accuracy is often higher when training on comparable amounts (costs) of easy rather than hard data. Here, we suppose hard data is 2x costlier to collect. Results are for Llama-2-70b with linear probes.}
   \label{fig:rq4_annotation_cost_appendix}
   \vspace{-1pt}
\end{figure}

\vspace{5pt}
\noindent \textbf{Design.} For (1) the budget assumption, we fit linear probes on either easy or hard data using datasets of sizes in \{10, 20, 40, 80, 160, 320\}.
We then show hard data test performance vs. training \emph{cost}, assuming hard data costs twice as much as easy data to collect, meaning labeling 40 easy training points is equivalent in cost to labeling 20 hard training points. 
For (2) the noise assumption, we assume that easy data is mislabeled $p\%$ of the time, while hard data is mislabeled $2p\%$ of the time. Here, we measure test performance on hard data given different values of $p$.
Note the 1:2 data collection cost ratio is almost exactly the ratio observed in MMLU-STEM-5, which contains 603 college level questions and 1143 high school questions, and a 1:2 labeling error ratio is plausible as well given expert human accuracy on datasets like MMLU (estimated at 89.8\%) and GPQA (estimated at $\leq$72\% for difficult graduate level STEM questions) \citep{hendrycks2020measuring, rein2023gpqa}. 

\vspace{4pt}
\looseness=-1
\noindent \textbf{Results.} We show results for the data budget assumption in Fig. \ref{fig:rq4_annotation_cost_appendix}. In terms of hard test accuracy, there is more often than not an advantage to fitting a model to easy data rather than hard data when the cost ratio between them is 1:2. 

\begin{figure}[!t]
   \centering
   \vspace{-1pt}
   \includegraphics[width=.46\textwidth]{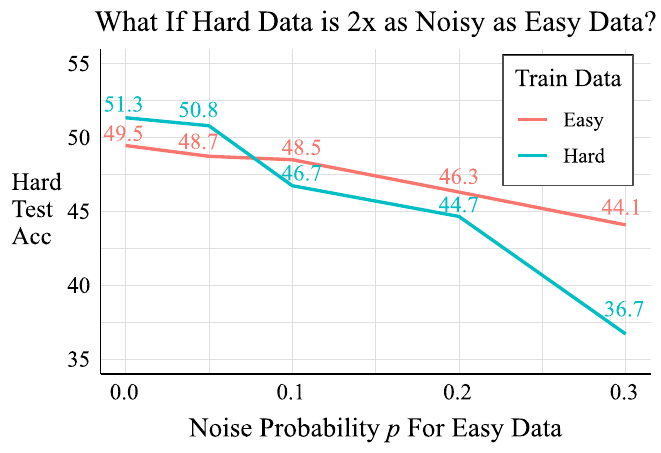}
   \vspace{-7pt}
   \caption{When hard data has noisier labels than easy data, finetuning on easy data can give better hard test performance (shown for MMLU-STEM-5 using Llama-2-70b with a linear probe).}
   \vspace{-9pt}
   \label{fig:rq4_noise_data_learned}
\end{figure}

For the noise assumption, we draw a similar conclusion based on the results for MMLU in Fig.~\ref{fig:rq4_noise_data_learned}. 
For instance, easy data is preferable when its labeling error rate is 10\% (or higher), meaning the error rate for hard data is 20\% (or higher). 
Since high error rates are possible for difficult domain questions \citep{rein2023gpqa}, there are plausible settings where it is better to finetune on easy data than hard data due to label noise.

\subsection{RQ4: Is Easy-To-Hard Generalization Consistent Across Model Scale and Train-Test Hardness Gap Size?}
\label{sec:RQ4}

\begin{figure*}[!ht]
   \centering
   \includegraphics[width=.97\textwidth]{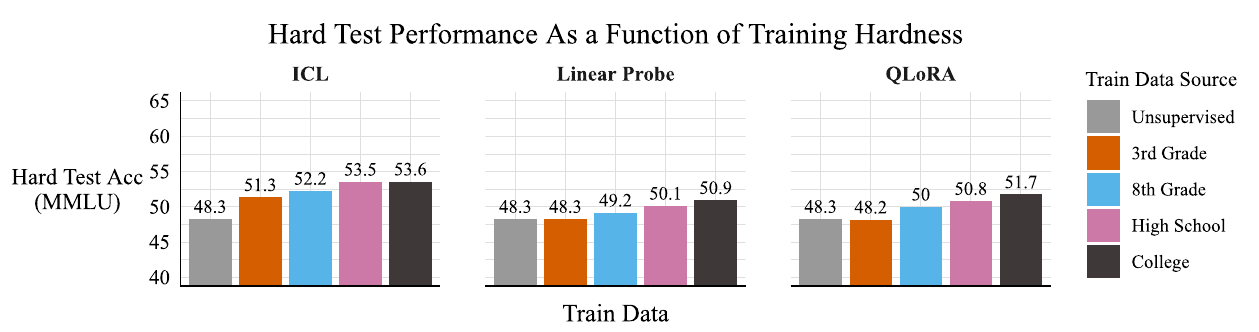}
   \vspace{-4pt}
   \caption{Performance on hard test data may begin to decline when finetuning on very easy data. Results shown for college-level STEM questions using Llama-2-70b (see other hardness measures, models in Appendix Figs. \ref{fig:rq5_additional_variables}, \ref{fig:rq5_model_robustness}).}
   \label{fig:RQ5_train-test-diff-third-grade-to-college}
   \vspace{-4pt}
\end{figure*}

We consider two questions likely to be relevant as models become more capable: (1) how does the Supervision Gap Recovered change as models scale in size, and (2) how does hard test performance change as the gap between train and test hardness grows? We are interested in these questions because in various settings AI performance may exceed human expert performance, and we want to know whether it will become more and more difficult to supervise models as this occurs. 

\vspace{5pt}
\noindent \textbf{Design.} 
For question (1), we measure easy-to-hard and hard-to-hard performance on MMLU for Llama-2 at three different model sizes: 7b, 13b, and 70b. For question (2), we test models on hard MMLU data (college STEM questions), while finetuning them on \emph{even easier} datasets than high school MMLU: 3rd and 8th grade ARC questions.

\looseness=-1
\vspace{6pt}
\noindent \textbf{Results.} First, we find that \textbf{models show similar levels of Supervision Gap Recovered across sizes} (Appendix Fig. \ref{fig:RQ5_mmlu_scaling}). SGR is near 100\% for model sizes between 7b and 70b.
For our second question, we find that the difference between train and test hardness may have an effect on test performance (Fig. \ref{fig:RQ5_train-test-diff-third-grade-to-college}). 
Across methods, we see some decline in generalization once the gap between train and test hardness becomes sufficiently large. However, even 3rd grade supervision can be surprisingly effective for college STEM questions (e.g. SGR falls from 74\% to 57\% when fitting to 8th grade vs. 3rd grade questions using ICL). 
In an evaluation across our other hardness measures, easy training data is only marginally worse than medium training data (see Appendix Fig. \ref{fig:rq5_additional_variables}). 
Together these results suggest that, while (1) easy supervision remains effective as models scale up, (2) easy-to-hard performance may begin to decline when the gap between train and test hardness becomes sufficiently large.

\section{Discussion}

\looseness=-1
\textbf{Are Our Tasks Hard Enough to Provide Generalizable Easy-To-Hard Results?}
Benchmark datasets for LMs now require specialized domain expertise \cite{hendrycks2020measuring}. 
The largest difficulty gap that we test in this paper is between 3rd grade and college level STEM questions (using ARC and MMLU). Concurrent work has called for work studying gaps as large as 3rd grade to 12th grade
\cite[described as ``huge leaps in generalization'';][]{burns2023weak}. Therefore, we see our results as relevant for future work that may operationalize ``easy'' and ``hard'' differently.

\looseness=-1
\vspace{5pt}
\noindent\textbf{How Do LMs Solve Hard Problems From As Few As Ten Easy Examples?} 
Our results suggest that finetuning on even small amounts of easy data successfully elicits relevant knowledge from LMs in a way that is largely invariant to datapoint hardness (we do not conclude that we are teaching LMs entirely new skills).
This could be because this kind of finetuning encourages models to answer questions based on ``truthfulness'' representations of text, which should be invariant across domain and data hardness \cite[see][]{marks2023geometry}.
We emphasize that we do not interpret our results as models merely ``learning the task format'' as opposed to true generalization: we also fit models using ICL prompts that are trivially simple but match the task format for MMLU and StrategyQA, and find that model performance varies based on prompt data hardness and not simply prompt task format (see Appendix Fig. \ref{fig:RQ3_task_format}). 
Hence it appears that fitting to easy data encourages models to give correct outputs for hard questions.

\section{Conclusion}

\looseness=-1
We study the problem of easy-to-hard generalization, showing that (1) several meaningful human and model-based hardness measures disagree about which data is hardest; (2) LMs trained on easy data often perform nearly as well as those trained on hard data, recovering 70-100\% of the Supervision Gap between an unsupervised lower bound and hard-to-hard upper bound; (3) practically, one can perform better on hard test data by collecting and training on easy data rather than hard data when the hard data is noisier or costlier to collect; and (4) SGR may begin to decline when the gap between train and test hardness becomes sufficiently large.
These results are robust across datasets, training methods, hardness measures, and model size.
Our findings suggest that the scalable oversight problem may be easier than previously thought.

\section*{Limitations}

We aim to study how models generalize from settings where humans can easily label data to those where humans have difficulty labeling data. Since we require ground truth labels to evaluate model generalization, this limits our evaluations to datasets where humans \emph{have} reliably labeled the data. While we aim to test a broad range of train and test difficulties (like 3rd grade training data and college-level test data), our results may not generalize to settings with difficulty levels besides those we test in this paper, especially test settings where almost no amount of human effort can reliably produce accurate data labels (like unsolved scientific questions). 

Additionally, we are not able to verify that test questions for ARC, MMLU, StrategyQA, and GSM8k are not in training datasets of the open models used in this paper, including Llama-2-70b and Mixtral 8x-7b. This means we cannot be certain that models are generalizing from easy to hard data, rather than reporting memorized hard question answers. While this concern currently affects all research on open source LLMs, including those with public datasets (since finding test set contamination in pretraining data is an unsolved problem), we note that it would be beneficial for future work to evaluate easy-to-hard generalization on test sets that are either private or collected after pretraining data collection cut-offs.

\section*{Ethics Statement}

We hope that positive results in easy-to-hard generalization imply that we can train models to perform well in niche domain like biology, chemistry, medicine, law, engineering, etc., without demanding significant amounts of expert time and spending large amounts of money in order to annotate data in these settings. In this way, we might make LLMs more useful for hard tasks while requiring less human effort to supervise their training. At the same time, we note that there are dual use and human labor displacement concerns around improving model capabilities in such domains. Ultimately we hope for LLMs to be deployed responsibly, so that they can be used to further human values and not for any ill intent. 

\bibliography{main}
\bibliographystyle{acl_natbib}

\appendix

\section{Measuring Minimum Description Length}
\label{app:MDL}

In addition to our human hardness measures, we employ a model-based metric based on minimum-description-length \cite{voita2020information, swayamdipta2020dataset, perez2021rissanen}. Since experiments use up to 70b parameter LMs, we measure MDL with models in the 7b parameter range, including Falcon-7b \cite{falcon}, Mistral-7b \cite{mistral}, Persimmon-8b \cite{persimmon}, and Llama-1-7b \cite{touvron2023llama}. To get one MDL per datapoint, we average the MDL scores obtained for each of the four models. For a single model, we obtain a score by training $N$ models on training sizes in $n \in \{5, 20, 80, 340, 900\}$ (roughly log-uniform) when fitting models with linear classifier heads or QLoRA and averaging model label confidences across these $n$ per-datapoint scores. For ICL we \textit{compute MDL using no training data}, i.e. $n=0$. In this way, our Linear Probe and QLoRA MDL metrics represent MDL according to the theoretical definition which involves increasing amounts of training data, while MDL (ZS Prompt) represents the confidence that 7b models assign to labels for data with no supervision. All metrics are then used for assessing how stronger models will perform on data that weaker models find to be hard. See distributions of MDL scores on each dataset in Fig. \ref{fig:hardness_var_distr}. We only measure probing-based and QLoRA-based MDL for ARC and GSM8k, where we have sufficient data to set aside $n=1000$ points (up to 900 for training and 100 for model validation).

\section{Additional Results}
\label{app:additional_results}

We include a number of additional results in this section. 

\begin{figure*}[!ht]
   \centering
   \includegraphics[width=.87\textwidth]{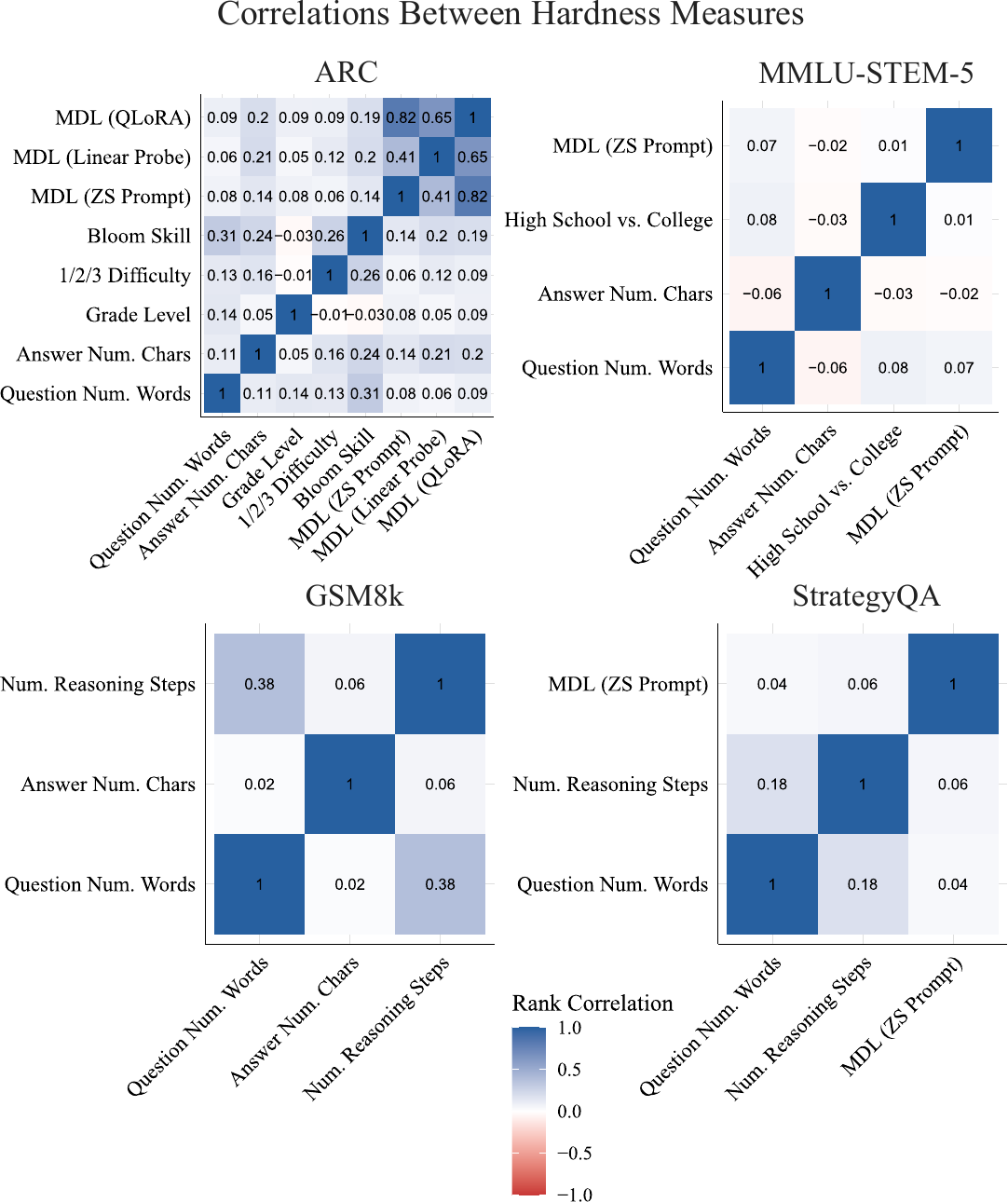}
   \vspace{-6pt}
   \caption{Correlations between hardness measures across datasets (Spearman rank order correlation). We omit MDL for GSM8k because 7b parameter models obtain extremely high loss on GSM8k problems, and MDL is valid as a metric only when using a reasonably good model of the data.}
   \label{fig:rq1_appendix}
\end{figure*}

\begin{figure*}[!ht]
   \centering
   \includegraphics[width=.91\textwidth]{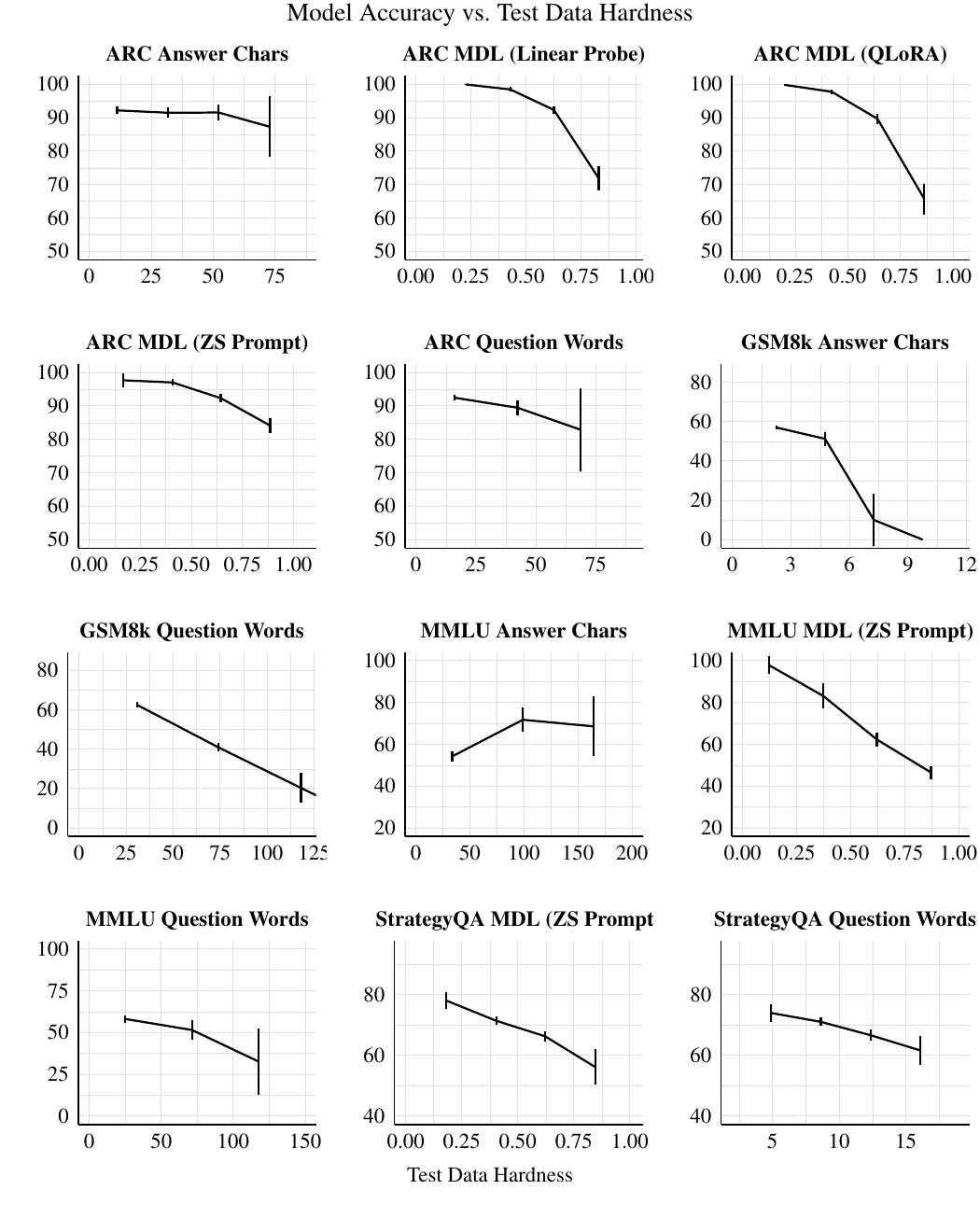}
   \vspace{-16pt}
   \caption{Test accuracy declines as test data hardness increases (shown for \textbf{additional hardness measures}),  with the exception of MMLU Answer Chars. Error bars are 95\% CIs showing test sample variance.}
   \label{fig:rq2_appendix}
\end{figure*}

\begin{figure*}[!ht]
   \centering
   \includegraphics[width=.91\textwidth]{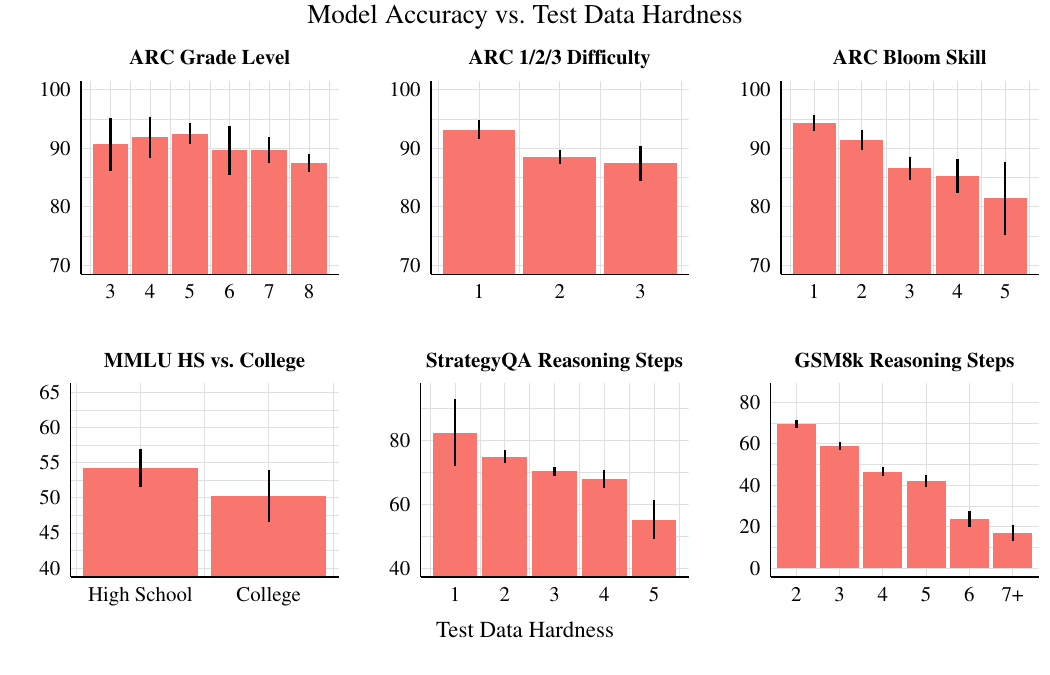}
   \vspace{-16pt}
   \caption{Test accuracy declines as test data hardness increases (shown for \textbf{QLoRA with Llama-2-70b}). Error bars are 95\% CIs showing test sample variance.}
   \label{fig:rq2_qlora}
\end{figure*}

\begin{figure*}[!ht]
   \centering
   \includegraphics[width=.91\textwidth]{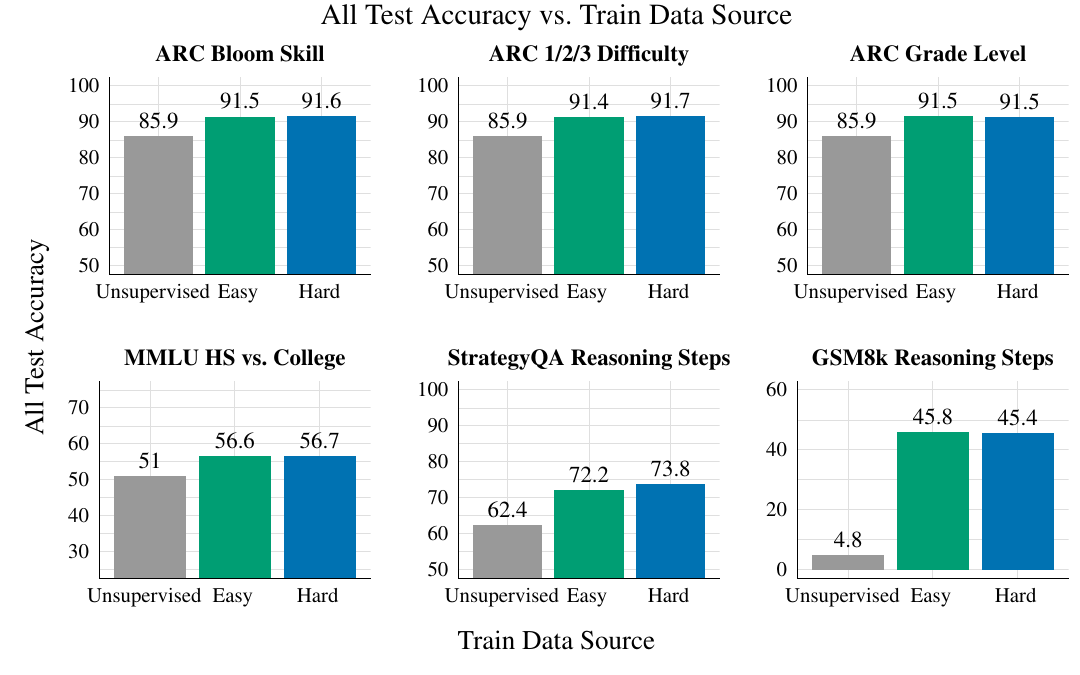}
   \vspace{-6pt}
   \caption{Easy-to-hard generalization measured on \textbf{all test data} (not subsetting to hard test data), while training on easy/hard data defined according to each hardness measure (using Llama-2-70b prompted with $k\leq10$ examples). Results are similar to testing on hard data, except for GSM8k, where accuracy on the whole data distribution becomes comparable (training on easy data outperforms hard data on easy/medium test data).}
   \label{fig:RQ3_all_acc}
\end{figure*}

\begin{figure*}[!ht]
   \centering
   \includegraphics[width=.91\textwidth]{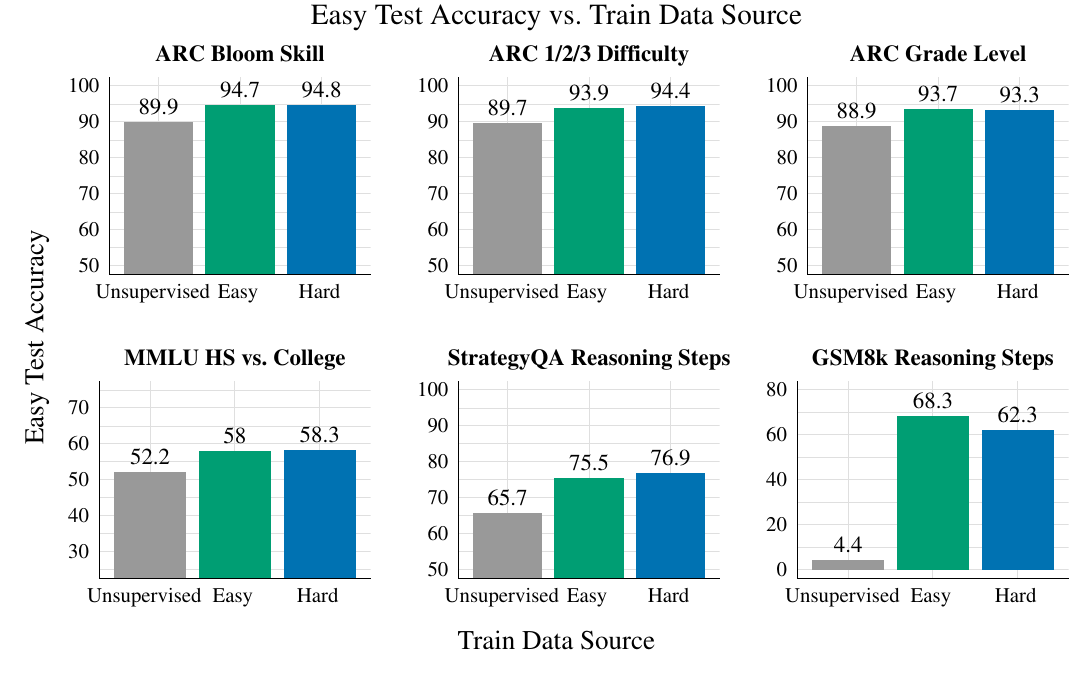}
   \vspace{-6pt}
   \caption{Easy-to-hard generalization measured on \textbf{easy test data}, while training on easy/hard data defined according to each hardness measure (using Llama-2-70b prompted with $k\leq10$ examples). This plot shows hard-to-easy generalization for each dataset, compared to easy-to-easy genearlization. On some datasets, hard data makes for better training data, while for others, easy training data is better for easy test performance.}
   \label{fig:RQ3_easy_acc}
\end{figure*}

\begin{figure*}[!ht]
   \centering
   \includegraphics[width=.99\textwidth]{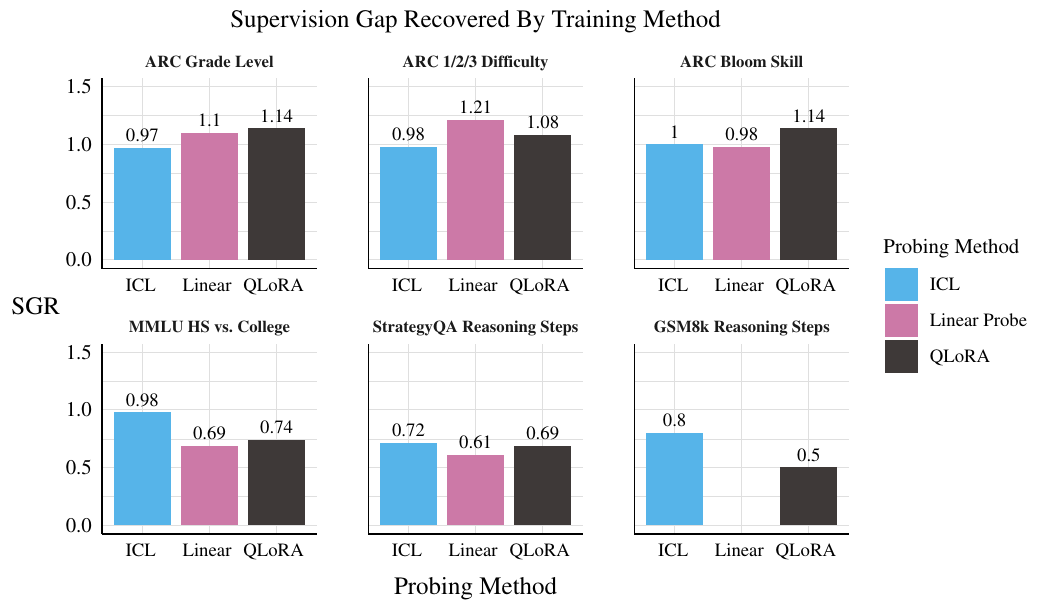}
   \vspace{-6pt}
   \caption{The Supervision Gap Recovered (SGR) shown by training method. Easy-to-hard generalization varies somewhat by training method used, but SGR remains surprisingly high across datasets for the two most effective training methods, ICL and QLoRA.}
   \label{fig:RQ3_SGR_by_methods_main}
\end{figure*}

\begin{figure*}[!ht]
   \centering
   \includegraphics[width=.91\textwidth]{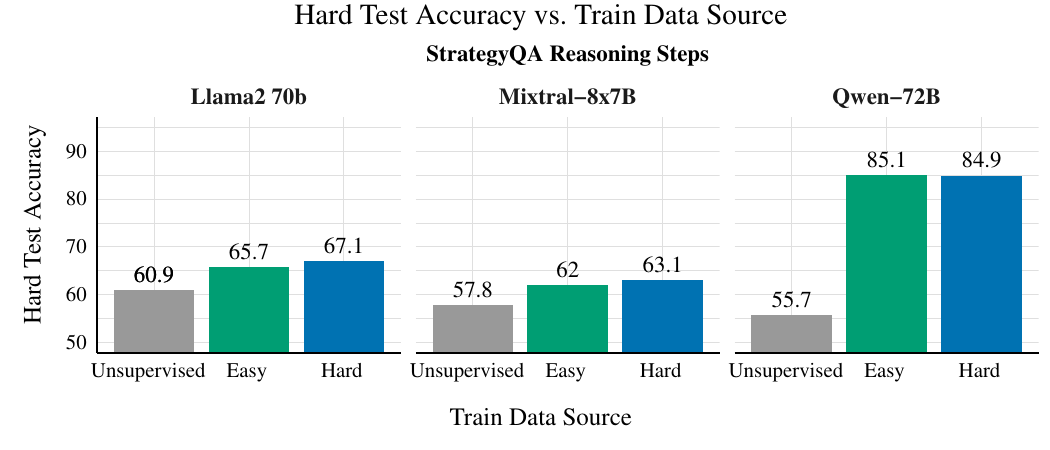}
   \vspace{-8pt}
   \caption{Easy-to-hard generalization results for different models on StrategyQA, using Llama-2-70b prompted with $k=4$ examples using CoT. Results are similar for Llama-2-70b and Mixtral, while Qwen appears to have been trained on StrategyQA data in a CoT format.}
   \label{fig:RQ3_strategyqa_model_robustness}
\end{figure*}

\begin{figure*}[!ht]
   \centering
   \includegraphics[width=.91\textwidth]{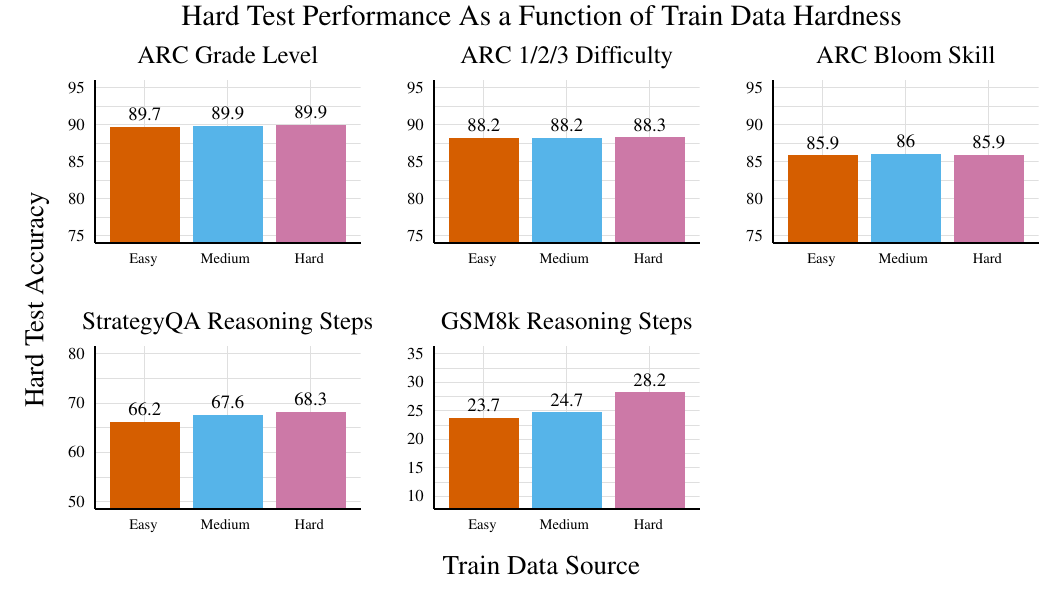}
   \vspace{-6pt}
   \caption{Test performance on hard data sometimes declines more significantly as the gap between train and test hardness grows, but often the difference between training on Medium and Easy data is relatively small in nature (using Llama-2-70b prompted with $k\leq10$ examples). MMLU not shown here since there are only two hardness levels for that dataset (high school vs. college). See Fig. \ref{fig:RQ5_train-test-diff-third-grade-to-college} for more results training on college vs. high school vs. 8th grade vs. 3rd grade data.}
   \label{fig:rq5_additional_variables}
\end{figure*}

\begin{figure*}[!ht]
   \centering
   \includegraphics[width=.91\textwidth]{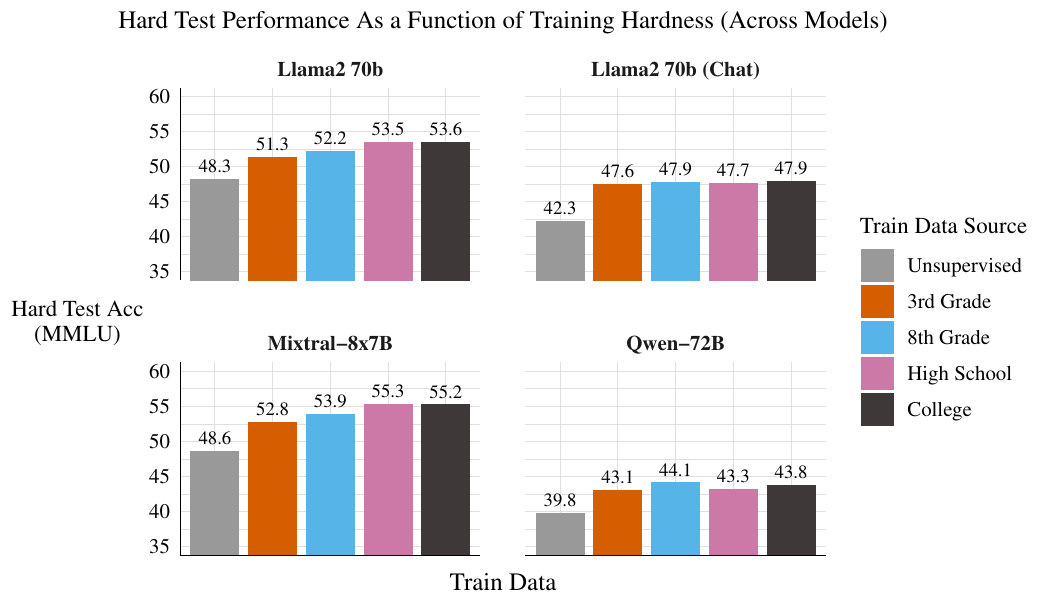}
   \vspace{-6pt}
   \caption{Test performance on hard data declines as the gap between train and test hardness grows for reasoning datasets, \textbf{across models}, using ICL with $k=10$.}
   \label{fig:rq5_model_robustness}
\end{figure*}

\begin{figure*}[!ht]
   \centering
   \includegraphics[width=.99\textwidth]{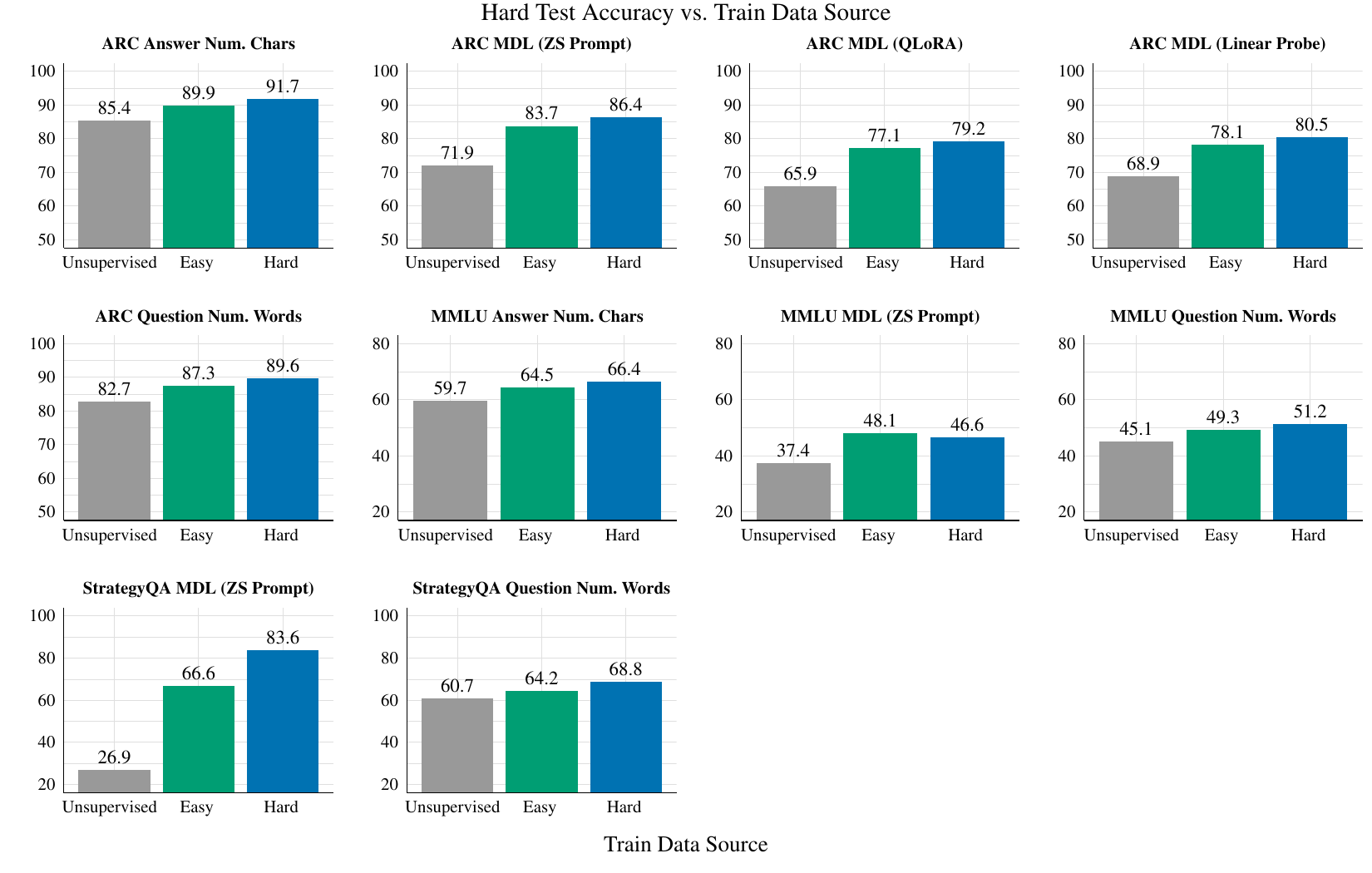}
   \vspace{-6pt}
   \caption{Easy-to-hard generalization for additional hardness measures for each dataset, using Llama-2-70b with ICL. SGR values remain high across possible hardness measures, meaning easy data provides surprisingly good supervision. We do not represent Answer Num Chars. for StrategyQA here because that would cleanly divide the data into `no' and `yes' categories. We do not conduct any additional experiments for GSM8k hardness measures as these experiments (involving sampling CoTs with $t=300$ tokens) are extremely computationally expensive.}
   \label{fig:RQ3_appendix}
\end{figure*}

\begin{figure*}[!ht]
   \centering
   \includegraphics[width=.99\textwidth]{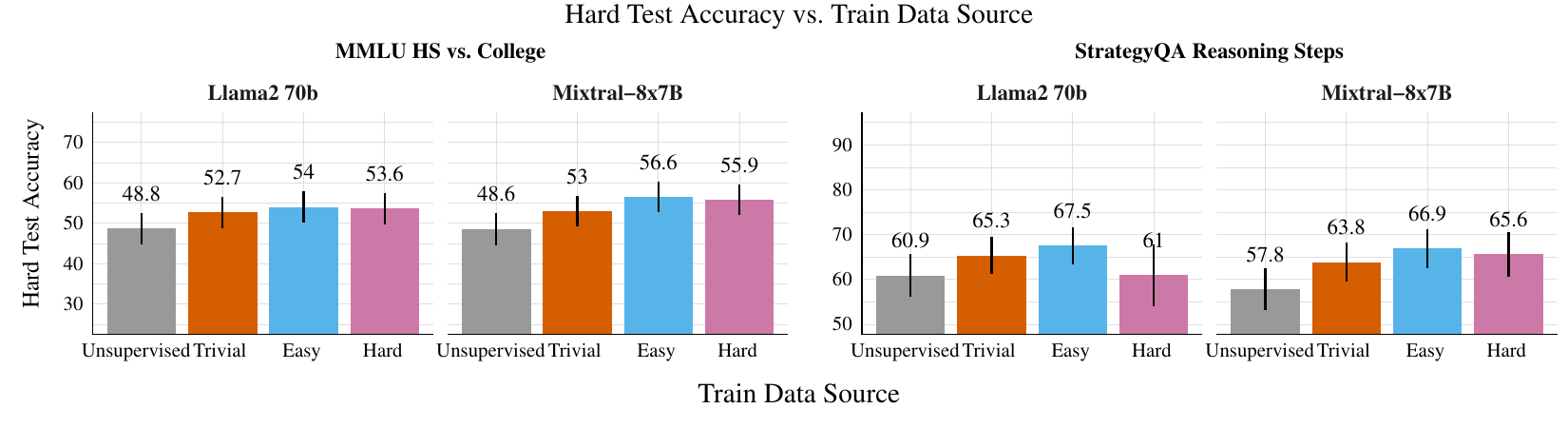}
   \vspace{-6pt}
   \caption{We consider prompts that contain trivially simple problems that match the task format for MMLU and StrategyQA (see \autoref{tab:task_format_prompts}). Results are shown for ICL without CoT, using Llama-2-70b and Mixtral-8x7b. We conclude that task format alone is not sufficient for encouraging generalization to hard data, because hard test performance varies by train data hardness, while all prompts share the task format (the best supervised prompt is better than other supervised prompts at a statistical significance threshold of $p<.05$ for three of four comparisons). For example, Llama-2-70b actually does not generalize at all from hard prompt examples for StrategyQA, suggesting that task format alone is not enough for generalization.}
   \label{fig:RQ3_task_format}
\end{figure*}

\begin{enumerate}[leftmargin=31pt, itemsep=-2pt]
    \item For a table of model accuracies by training method in an all-to-all setup, see Table \ref{tab:model_acc_table}.
    \item We show correlations between hardness measures for all data in Fig. \ref{fig:rq1_appendix}.
    \item We show that test accuracy declines with test data hardness for QLoRA in Fig. \ref{fig:rq2_qlora}, and using ICL with additional hardness measures in Fig. \ref{fig:rq2_appendix}.
    \item We show easy-to-hard generalization as measured on \emph{all} test data (not subsetting to hard test data) in Fig. \ref{fig:RQ3_all_acc}, as well as testing on \emph{easy} test data in Fig. \ref{fig:RQ3_easy_acc}. 
    \item We show easy-to-hard generalization on additional hardness measures in Fig. \ref{fig:RQ3_appendix}.
    \item We show SGR statistics for all training methods in Fig. \ref{fig:RQ3_SGR_by_methods_main}.
    \item We give SGR estimates along with confidence intervals and $p$-values, obtained by block bootstrap, in Table \ref{tab:RQ3-p-values}.
    \item We show easy-to-hard generalization on StrategyQA across different models in Fig.~\ref{fig:RQ3_strategyqa_model_robustness}.
    \item We give additional results for test performance by multiple training difficulty levels in Fig.~\ref{fig:rq5_additional_variables}.
    \item We give test performance by training difficulty for multiple $\sim$70b models in Fig. \ref{fig:rq5_model_robustness}.
    \item In Fig. \ref{fig:RQ3_task_format}, we test the hypothesis that task format alone is taught by training data, as opposed to true generalization. We list examples used for ``task format only'' prompts, which are trivially simple examples matching multiple-choice or yes/no answer prompts in our data. 
\end{enumerate}

\begin{figure*}[!ht]
   \centering
   \includegraphics[width=1\textwidth]{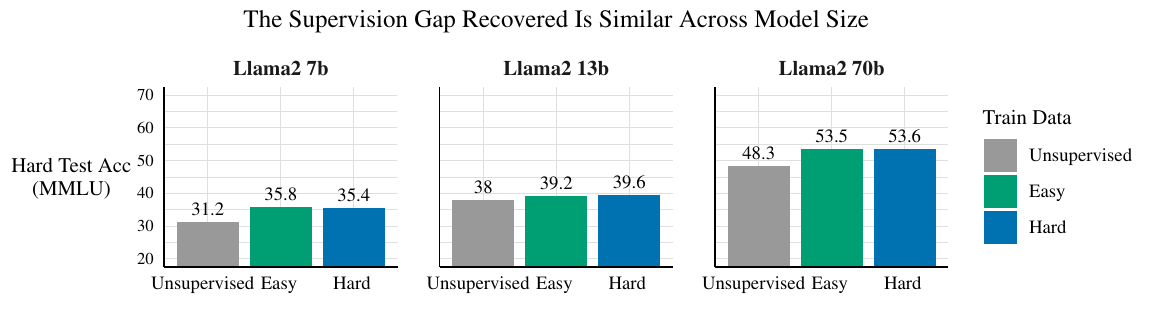}
   \vspace{-28pt}
   \caption{Models perform as well on hard MMLU data when prompted with easy MMLU data as they do when prompted with hard data, regardless of model size ($k=10$ examples used for ICL).}
   \label{fig:RQ5_mmlu_scaling}
\end{figure*}

\begin{table*}[ht]
\centering
\small
\begin{tabular}{lllllr}
\toprule
Dataname    & Hardness Measure & SGR Estimate        & Test Hardness & $n$    \\ 
\midrule
ARC         & Grade Level         & 0.96 ± 0.10 ($p$ < 1e-4) & Hard     & 1588    \\
ARC         & 1/2/3 Difficulty    & 0.98 ± 0.36 ($p$ = 0.0033) & Hard     & 1588    \\
ARC         & Bloom Skill         & 1.00 ± 0.20 ($p$ < 1e-4) & Hard     & 1588    \\
MMLU        & HS vs. College      & 0.97 ± 0.59 ($p$ = 0.0158) & Hard     & 603     \\
StrategyQA  & Num Reasoning Steps & 0.72 ± 0.93 ($p$ = 0.0788) & Hard     & 427     \\
GSM8k       & Num Reasoning Steps & 0.79 ± 0.60 ($p$ = 0.0125) & Hard     & 333     \\
\midrule
ARC         & Grade Level         & 1.00 ± 0.09 ($p$ < 1e-4) & All      & 3521    \\
ARC         & 1/2/3 Difficulty    & 0.96 ± 0.08 ($p$ < 1e-4) & All      & 3521    \\
ARC         & Bloom Skill         & 0.98 ± 0.08 ($p$ < 1e-4) & All      & 3521    \\
MMLU        & HS vs. College      & 1.00 ± 0.27 ($p$ = 0.0001) & All      & 1746    \\
StrategyQA  & Num Reasoning Steps & 0.87 ± 0.32 ($p$ < 1e-4) & All      & 2290    \\
GSM8k       & Num Reasoning Steps & 0.98 ± 0.39 ($p$ = 0.0003) & All      & 2065    \\ \bottomrule
\end{tabular}
\caption{Supervision Gap Recovered (SGR) statistics for Llama-2-70b with ICL, on hard test data or all test data, defined per dataset and hardness measure. Confidence intervals are 95\% CIs estimated by block bootstrap (accounting for test data variance and train data variance), and $p$-values represent a test for a difference from 0.}
\label{tab:RQ3-p-values}
\end{table*}

\section{Dataset Details}
\label{app:dataset_details}

\begin{figure*}[!ht]
   \centering
   \includegraphics[width=.91\textwidth]{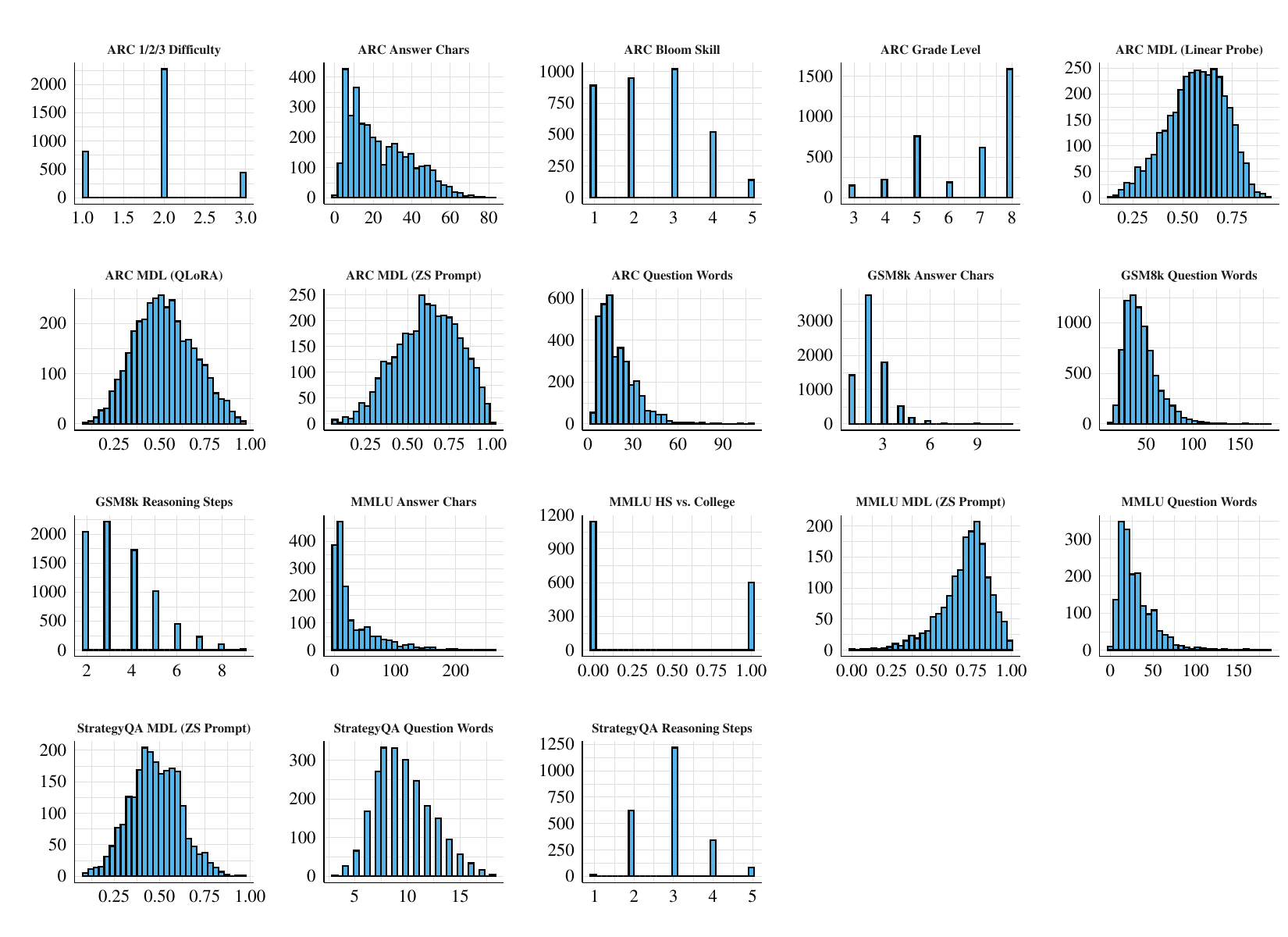}
   \vspace{-6pt}
   \caption{Distributions for hardness measures for each dataset and hardness measure.}
   \label{fig:hardness_var_distr}
\end{figure*}

We provide additional details for each dataset below, including test data subsetting decisions (see final sample sizes in Table \ref{tab:RQ3-p-values}). All datasets are publicly available, and license information is included via the links provided. All datapoints are in English. See also Table \ref{tab:hardness_vars} for a list of which hardness measures are available for which datasets, as well as sample sizes. See Fig. \ref{fig:hardness_var_distr} for histograms of hardness measures per dataset.

\begin{enumerate}[leftmargin=27pt, itemsep=-1pt, label=$\bullet$]
\item \textbf{ARC} \cite{Clark2018ThinkYH}:\footnote{\url{https://huggingface.co/datasets/ai2_arc}} U.S. gradeschool science questions in multiple-choice format. We combine the ARC-Easy and ARC-Challenge splits. We release metadata including human hardness metadata accompanying the original source of the questions in our codebase at \url{https://github.com/allenai/easy-to-hard-generalization}. We set aside 1000 points for MDL computation, so experiments are conducted on $n=3521$ test points.
\item \textbf{MMLU} \cite{hendrycks2020measuring}:\footnote{\url{https://huggingface.co/datasets/tasksource/mmlu}} Domain-specific multiple-choice questions for a large number of domains. We subset to high school and college level math, physics, biology, chemistry, and computer science questions (termed MMLU-STEM-5). Here, grade level is high school (HS) vs. college. See \autoref{fig:examples-fig} (left) for an example.
\item \textbf{StrategyQA} \cite{Geva2021DidAU}:\footnote{\url{https://huggingface.co/datasets/wics/strategy-qa}} General knowledge trivia questions requiring compositional reasoning over individual facts. The number of facts that must be combined forms the ``Num. Reasoning Steps'' measure.
\item \textbf{GSM8k} \cite{cobbe2021training}:\footnote{\url{https://huggingface.co/datasets/gsm8k}} U.S. grade school math word problems. The number of steps in the solution to the problem forms the ``Num. Reasoning Steps'' measure and is obtained from the human-annotated reasoning chain collected for each problem. We set aside 1000 points for MDL computation, then further subset to about $n=2000$ test points given the extreme expense of sampling CoTs with $t=300$ tokens for 70b parameter models. 
See \autoref{fig:examples-fig} (right) for an example datapoint.
\end{enumerate}

Distributions for hardness measures for each dataset and hardness measure (from Table. \ref{tab:hardness_vars}) are shown in Fig. \ref{fig:hardness_var_distr}.

\section{Modeling and Tuning Details}
\label{app:tuning_details}

We provide additional information around model tuning for each training method here. 

\noindent\textbf{GPU Cost.} We run experiments on NVIDIA A6000 GPUs. A typical experiment setting is running Llama-2-70b over $n=2000$ datapoints, quantized 8bit, with ICL with $k=8$, with batch size 1, using 5 random seeds for prompt data selection. Such an experiment requires 4 GPUs and takes about 10 hours to complete. Experiment with CoT are more expensive, taking about one hour per $n=200$ test points on GSM8k, using a decoding length of $t=300$ tokens. Linear probing experiments have negligble runtime as we save the hidden states to file, avoiding the need to rerun model forward passes, while QLoRA experiments take about as long as ICL experiments (slightly faster for CoT settings). 

\noindent\textbf{Prompt Templates.} All training methods use the same prompts, one per dataset, that were selected based on their success in past work \cite{saha2023can}. We show prompts for ARC and MMLU in Table \ref{tab:prompt_formats}.
In this template, the \{\} placeholders are filled with the question, four answer choices, and a single answer choice (explained next). We use this prompt for multiple choice scoring of the four answer choices \emph{for all methods}, meaning that we run four forward passes to either (1) compute answer choice probabilities for each answer choice for ICL and QLoRA, or (2) collect final answer choice token representations for each answer choice for linear probing. Thus the final answer slot, ``A: \{\}'' is populated by each answer choice once. 
When prompting with $k$ in-context examples for ICL, we separate examples with a line break (one line between each pair of examples).

For StrategyQA and GSM8k, we use a different prompt format for CoT, shown in Table \ref{tab:prompt_formats}. 
In this template, the \{\} placeholders are filled with the question, the human reasoning chain, and the answer choice (\textit{only for in-context examples}). This prompt is used to generate new reasoning chains and answers at test time, so there is no text included after ``A:'' for the test input.
When prompting with $k$ in-context examples for ICL, we separate examples with a line break (one line between each pair of examples). The exception to this formatting is for GSM8k's Unsupervised Baseline, which uses a ``Let's think step-by-step'' prompt (we also considered this for StrategyQA, but zero-shot answer choice scoring worked better). The step-by-step prompt is shown in Table \ref{tab:prompt_formats}. The test input is supplied to the curly brace placeholder. 

\begin{table}
    \setlength{\tabcolsep}{11pt}
    \small
    \centering
    \begin{tabular}{l}
        \toprule
        \vspace{2pt}
        \textbf{ARC+MMLU prompt} \\
        \hline
        Question: \{\} \\
        A) \{\} \\
        B) \{\} \\
        C) \{\} \\
        D) \{\} \\
        Answer: \{\} \\
        \hline
        \vspace{2pt}
        \textbf{StrategyQA+GSM8k prompt} \\
        \hline
        Q: \{\} \\
        A: \{\} So the answer is \{\}
        \vspace{3pt}
        \\
        \hline
        \vspace{2pt}
        \textbf{Unsupervised GSM8k prompt} \\
        \hline
        \vspace{2pt}
        Q: [question text here] \\
        \vspace{2pt}
        A: Let's think step by step. \\
        1. [step one] \\
        2. [step two] \\
        \ldots \\
        N. [last step] \\
        Therefore, the answer is [answer here]. \\
        Now you try! \\
        \vspace{2pt}
        Q: \{\} \\
        \vspace{2pt}
        A: Let's think step by step. \\
        1. \\
        \bottomrule
    \end{tabular}
    \caption{Prompt formats used in this paper. Question text, reasoning text, answer choices, and answer text are imputed in curly brackets. The notation ``[step one]'' is literal, and no variables are imputed in these brackets. When in-context examples are included in the prompt, we separate each example with one empty line.}
\label{tab:prompt_formats}
\end{table}

\noindent\textbf{In-context Learning.} For ICL, we select $k=10$ for ARC and MMLU and $k=8$ for StrategyQA and GSM8k as we see diminishing returns to accuracy from larger values of $k$, and using larger $k$ values significantly slows down experiments.

\noindent\textbf{Linear Probing.} For Linear Probing, we fit a linear classifier to frozen LM hidden states. 
For a given question, we compute one representation per answer choice by concatenating the question and answer choice and feeding it to the model. To get a single representation from the LM forward pass, we concatenate the representations at the last token index (i.e., the last answer token) from the middlemost and last layer.
Then, we score each question-answer choice representation $z$ by applying the linear probe: $f(z;w) = w^Tz$. The answer choice with the highest score is returned as the prediction. The probe weight $w$ is trained using SGD to minimize cross-entropy loss on a dataset of frozen representations $Z = \{\{z_{i,j}\}_{j=1}^{j=|A|}\}_{i=1}^{N}$ for a dataset of $N$ training datapoints and $|A|$ answer choices. We optimize the weight with the default \texttt{SGD} implementation in PyTorch \cite{paszke2019pytorch} for 100 epochs, without early stopping on any dev data. Based on tuning experiments with Llama-2-13b on ARC, we chose to use SGD rather than AdamW, selecting the number of epochs for convergence, and we chose to use the middlemost and last layer representations (concatenated) rather than either on its own. Note this produces very high dimensional inputs, but by using $\ell_2$ regularization with $\lambda=1$, we are able to stably fit probes to these $2d$-dimensional input representations (where $d=8192$ for Llama-2-70b) with as few as $n=10$ training points. The learning rate was fixed at 5e-2. 

\noindent\textbf{Model Finetuning with QLoRA.} For QLoRA, we selected hyperparameters based on early experiments with Llama-13b on ARC. Based on this setting, we selected an adaptor rank of $r=16$ rather than $r=8$, with default hyperparameters otherwise, including default selected layers to optimize. We use AdamW with a a learning rate fixed at 1e-4, and the model is optimized with a batch size of 50. At least 10 gradient updates are performed, or a minimum of three epochs, whichever yields more gradient updates. This means that for train $n=160$ points and a batch size of 50, we generally perform 12 gradient update steps (3 epochs) in our experiments. 

\noindent\textbf{Decoding Steps.} To select the number of decoding steps for each datasets ($t=100$ for StrategyQA and $t=300$ for GSM8k), we wanted to make sure that we were generating reasoning chains long enough for Llama-2-70b to solve hard test questions. Therefore, we intentionally selected this parameter based on hard test performance, in order to use as small a valuable as possible (based on experiment efficiency) that did not compromise test performance on hard data.

\noindent\textbf{Quantization.} All models are run in 8bit quantization, except for Qwen and Mixtral, which are run in 16bit format, and falcon-7b and persimmon-8b, which are run in full precision. We observe no performance loss from quantization in any experiment.

\noindent\textbf{Training Size Controls.} For all experiments with linear probing and QLoRA, we use $n=160$ train points. While we would prefer to use more training data, the bottleneck we face is that fairly comparing easy-to-hard with hard-to-hard generalization requires both fixing the amount of training data and leaving enough hard data left over for testing. Since we have as few as $n=603$ hard test points for MMLU, we have to limit training data to $n=160$ points to leave enough test data for reasonably small confidence intervals.
Linear probing and QLoRA demonstrate good sample-efficiency when applied to Llama-2-70b, so we are able to obtain comparable (and sometimes better) performance than ICL across datasets using these methods.

\begin{table}[t]\setlength{\tabcolsep}{4pt}
\centering
\small
\begin{tabular}{l l l r r}
\hline
Dataset & Method & CoT & $n$ & Acc (\%) \\
\hline
ARC & ICL & No & 0 & 85.94 \\
ARC & ICL & No & 10 & 91.77 \\
ARC & Linear Probe & No & 160 & 89.48 \\
ARC & QLoRA & No & 160 & 89.47 \\
GSM8k & ICL & Yes & 0 & 4.80 \\
GSM8k & ICL & Yes & 8 & 56.24 \\
GSM8k & QLoRA & Yes & 160 & 52.64 \\
MMLU-STEM-5 & ICL & No & 0 & 48.30 \\
MMLU-STEM-5 & ICL & No & 10 & 56.83 \\
MMLU-STEM-5 & Linear Probe & No & 160 & 53.01 \\
MMLU-STEM-5 & QLoRA & No & 160 & 52.77 \\
StrategyQA & ICL & No & 0 & 62.40 \\
StrategyQA & ICL & No & 8 & 70.04 \\
StrategyQA & ICL & Yes & 8 & 72.86 \\
StrategyQA & Linear Probe & No & 160 & 68.79 \\
StrategyQA & QLoRA & No & 160 & 66.36 \\
StrategyQA & QLoRA & Yes & 160 & 75.09 \\
\hline
\end{tabular}
\caption{Model accuracy table when finetuned on randomly selected data from the whole data distribution and tested on the whole data distribution (zero-shot ICL rows included), using Llama-2-70b. Averaged over 5 seeds. Compare to Fig. \ref{fig:RQ3_all_acc}.}
\label{tab:model_acc_table}
\end{table}

\section{Statistical Testing}
\label{app:statistical_testing}

Here, we describe in greater detail how our statistical testing works. We aim to make the most of the data we have, e.g. $n=603$ hard datapoints for MMLU-STEM-5. Ultimately, we want to use a block bootstrap that resamples (1) test datapoints and (2) models (equivalent to picking which random seed was chosen for training), in order to account for variance due to limited test data, selection of training data, and any random training dynamics. Therefore, we run five random seeds for each experiment, randomly selecting training data and using all remaining data as test data. Each experiment produces a matrix (block) of results, with up to five model predictions per datapoint. Running this matrix through a block bootstrap that resamples rows and columns produces a confidence interval for the statistic of interest \cite{efron1994introduction}. When computing the mean of a resampled matrix, we ignore missing values (which represent that a datapoint was used for training and could not be tested on). We take 100,000 resamples. 

We can use a bootstrap to obtain estimates for our SGR statistic too. We aim to estimate the expected value of the random variable 
\begin{align*}
    \vspace{-1pt}
    \frac{\textrm{Easy} - \textrm{Unsupervised}}{\textrm{Hard} - \textrm{Unsupervised}}
    \vspace{-1pt}
\end{align*}
using the samples Unsupervised, Easy, and Hard representing their respective experiment outputs ($n \times 5$ matrices of model predictions). We perform a paired test with respect to test datapoints (resampling the same rows across each matrix), while not assuming random seeds are paired (resampling different columns for each matrix). Note that for the Unsupervised matrix, each column is identical because there is no prompt data. The results of this analysis are given in Table \ref{tab:RQ3-p-values}. When showing results for hard test data, we subset to just points that are hard according to their respective hardness measure. 

The one exception to this kind of bootstrap sampling is for results in Fig. \ref{fig:rq2_acc_vs_test_hardness} and \ref{fig:rq2_appendix}, where we only show CIs derived from test sample variance. In these plots, we average over five random training seeds. 

\begin{table}
    \setlength{\tabcolsep}{11pt}
    \small
    \centering
    \begin{tabular}{l}
        \toprule
        \vspace{2pt}
        \textbf{MMLU prompt} \\
        \hline
        Question: Which one is usually yellow? \\
A) Cheese \\
B) Apple \\
C) Carrot \\
D) Bread \\
Answer: Cheese \\
 \\
Question: What do we use to wash dishes? \\
A) Broom \\
B) Shovel \\
C) Soap \\
D) Hammer \\
Answer: Soap \\
 \\
Question: What color is the sky when it's sunny? \\
A) Grey \\
B) Blue \\
C) Black \\
D) Orange \\
Answer: Blue\\
...
        \vspace{3pt}
        \\
        \hline
        \vspace{2pt}
        \textbf{StrategyQA prompt} \\
        Q: Does cheese come from a plant?\\
A: no\\
\\
Q: Can you use a broom to sweep the floor?\\
A: yes\\
\\
Q: Is the sky green when it's sunny?\\
A: no\\
...\\
        \bottomrule
    \end{tabular}
    \caption{Example questions used for testing model performance based on task-format-only prompts. We use ChatGPT to generate 50 such questions for MMLU and 40 for StrategyQA. These are used for experiments using $k=10$ examples for MMLU and $k=8$ for StrategyQA, averaging over five different prompts (results shown in Fig. \ref{fig:RQ3_task_format}).}
\label{tab:task_format_prompts}
\end{table}

\end{document}